\pdfoutput=1

\documentclass[11pt]{article}

\usepackage[final]{acl}

\usepackage{times}
\usepackage{latexsym}
\usepackage{lipsum}  
\usepackage[normalem]{ulem}
\usepackage{subfigure}
\usepackage{subcaption}
\usepackage{graphicx}
\usepackage{multirow}
\usepackage{multicol}
\usepackage{adjustbox}
\usepackage{xcolor}

\usepackage[T1]{fontenc}

\usepackage[utf8]{inputenc}

\usepackage{microtype}

\usepackage{inconsolata}

\usepackage{url}            
\usepackage{hyperref}       
\usepackage{booktabs}       
\usepackage{amsfonts}       
\usepackage{nicefrac}       
\usepackage{microtype}      
\usepackage{xcolor}         
\usepackage{enumitem}
\usepackage{graphicx}
\usepackage{soul}

\usepackage{titletoc}

\newcommand{\modelName}{\textbf{\textsc{PoseStitch-SLT}}}

\newcommand{\BPCCISL}{BPCC-ISL}
\newcommand{\BPCCASL}{BPCC-ASL}
\newcommand{\BLIMPISL}{BLIMP-ISL}
\newcommand{\BLIMPASL}{BLIMP-ASL}

\title{\modelName: Linguistically Inspired Pose-Stitching for End-to-End Sign Language Translation}

\author{{\bf Abhinav Joshi}\thanks{\ \ Equal Contributions} \qquad
{\bf Vaibhav Sharma}\footnotemark[1] \qquad
{\bf Sanjeet Singh}\footnotemark[1] \qquad
 {\bf Ashutosh Modi} \\ 
 Department of Computer Science and Engineering\\
        Indian Institute of Technology Kanpur (IIT Kanpur) \\
  \texttt{\{ajoshi, svaibhav, sanjeet, ashutoshm\}@cse.iitk.ac.in}  
}

\begin{document}
\maketitle
\begin{abstract}

Sign language translation remains a challenging task due to the scarcity of large-scale, sentence-aligned datasets. Prior arts have focused on various feature extraction and architectural changes to support neural machine translation for sign languages. We propose \modelName, a novel pre-training scheme that is inspired by linguistic-templates-based sentence generation technique.  
With translation comparison on two sign language datasets, How2Sign and iSign, we show that a simple transformer-based encoder-decoder architecture outperforms the prior art when considering template-generated sentence pairs in training. We achieve BLEU-4 score improvements from 1.97 to 4.56 on How2Sign and from 0.55 to 3.43 on iSign, surpassing prior state-of-the-art methods for pose-based gloss-free translation. The results demonstrate the effectiveness of template-driven synthetic supervision in low-resource sign language settings. 

\end{abstract}

\vspace{-2mm}
\section{Introduction} \label{sec:intro}
\vspace{-2mm}

\begin{figure*}[h]
  \centering
  {
    \includegraphics[width=0.90\linewidth]{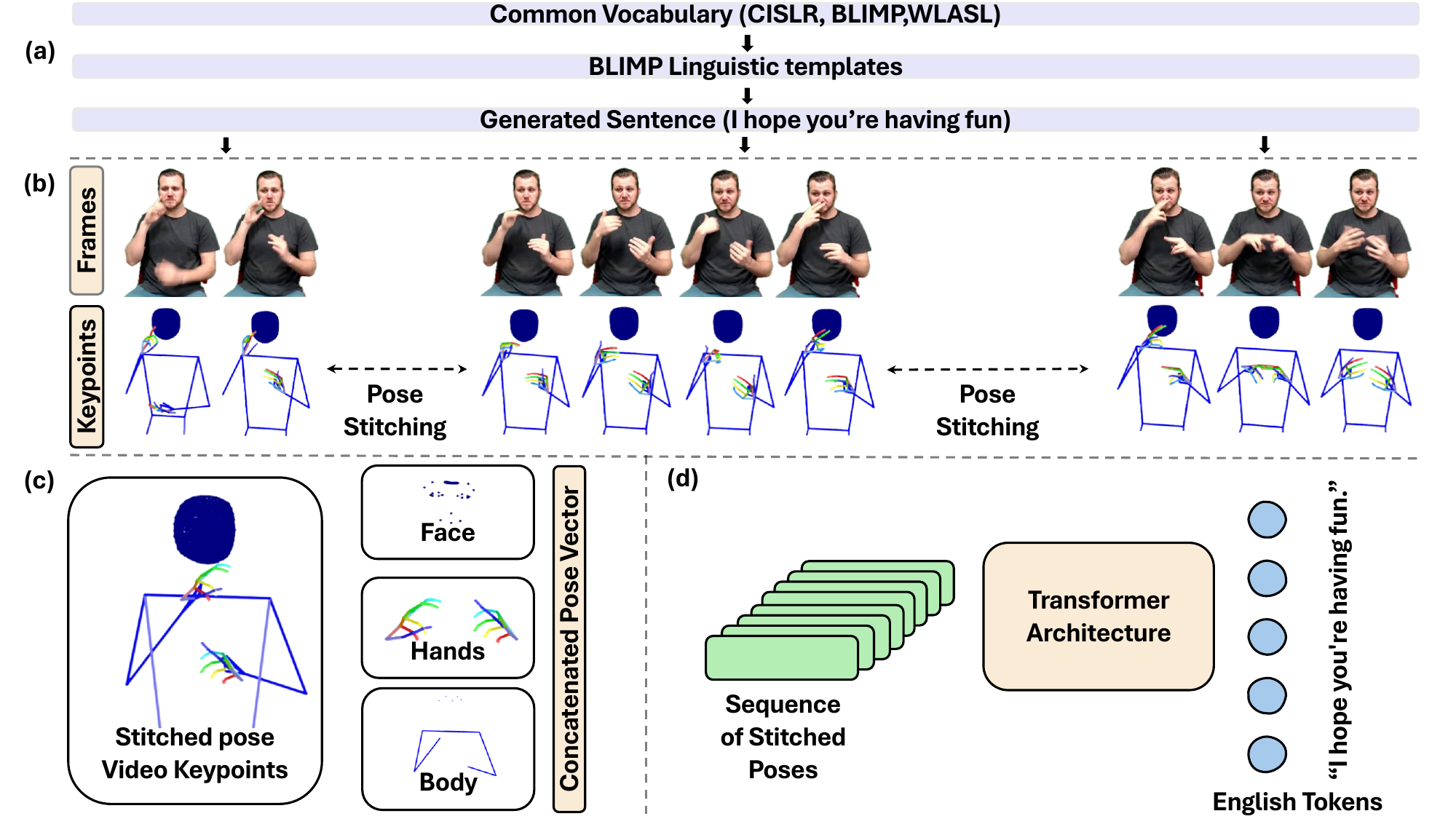}
  }
\caption{The figure shows the \modelName\ pipeline for generating a Sign Language Dataset for translation using pose stitching based on linguistic templates. Starting with a common vocabulary shared across datasets like CISLR, BLIMP, and WLASL, sentences are generated using BLIMP linguistic templates \cite{warstadt-etal-2020-blimp-benchmark}. For each generated sentence, corresponding sign language poses for words/gloss are stitched together. Frames are extracted from the stitched pose video to form a sequence of keypoints. These keypoints extracted from the face, hands, and body are concatenated to create pose vectors. This sequence of pose vectors serves as input to an encoder-decoder-based transformer model. The sentence example is taken from the How2Sign dataset, saying, “I hope you're having fun”.}
\label{fig:model-architecture}
\vspace{-5mm}
\end{figure*}

Sign languages are the primary mode of communication for over 70 million people from the deaf and hard of hearing (DHH) community globally, according to the World Federation of the Deaf \cite{10628563}. Despite increasing progress in natural language processing (NLP) \cite{wang2019superglue}, sign language processing remains significantly underexplored, both in terms of benchmark datasets and model development \cite{min2023recent, yin-etal-2021-including, moryossef2020real, jiang2024signclipconnectingtextsign}.
In contrast to spoken languages, sign languages are visual-gestural and multimodal, combining manual signs (e.g., hand shapes and movements) with non-manual cues (e.g., facial expressions and body posture) \cite{cd44457ffd6240d5a8c337b74b1ea267}. 
%
Moreover, training on raw sign videos raises fairness and privacy concerns due to signer-identifiable features. Pose-based approaches \cite{ko2019neural, camgoz2020multi}, which use 2D/3D keypoints from the face, hands, and body, provide a privacy-preserving alternative while retaining essential communication information.

\noindent In this work, we focus on pose-based, gloss-free Sign Language Translation (SLT) under real-world constraints of data scarcity and signer privacy. To address the lack of large-scale parallel data, we propose \modelName:  a \textbf{linguistically inspired pretraining} strategy that constructs synthetic pose-based sentence data, making use of publicly available word-level sign language datasets and linguistic templates \cite{warstadt-etal-2020-blimp-benchmark} to generate millions of grammatically diverse sentences in English. 
%
%
We use vocabularies from CISLR \citep{joshi-etal-2022-cislr} for Indian Sign Language (ISL) and WLASL for American Sign Language (ASL) \textcolor{blue}{ \citep{li2020word}}, two of the largest publicly available word-level sign datasets, covering approximately 4.5K and 2K words, respectively. While these vocabularies remain limited compared to spoken-language corpora, they represent the most extensive resources currently accessible to the SLT research community. This choice was made intentionally to enable scalable pretraining without relying on proprietary datasets or gloss annotations. Furthermore, our framework is modular and adaptable, designed to make the most of available resources while remaining extensible as new sign language datasets become available.
Fig.\ref{fig:model-architecture} illustrates the entire pipeline. 
Importantly, we employ a standard transformer architecture \cite{NIPS2017_3f5ee243} by design, in order to isolate and rigorously assess the impact of the proposed pretraining strategy. This design choice ensures that performance improvements are not confounded by model-specific enhancements, and highlights the generality of our approach. 
We evaluate our method on two challenging, publicly available benchmarks suitable for gloss-free, pose-only translation: How2Sign (ASL) \cite{Duarte_CVPR2021} and iSign (ISL) \cite{joshi-etal-2024-isign}. 
In a nutshell, we make the following contributions:
\begin{itemize}[noitemsep,nosep,leftmargin=*]
    \item We introduce \modelName, a \textbf{\textit{linguistically grounded strategy}} for synthetic dataset creation and model pretraining in \textbf{\textit{pose-based, gloss-free sign language translation.}}
    \item We demonstrate \textit{\textbf{state-of-the-art results}}, with BLEU-4 gains of \textit{\textbf{1.97–4.56 on How2Sign}} and \textit{\textbf{0.55–3.43 on iSign}}, using only pose inputs in a gloss-free setting and a standard Transformer architecture.
    \item We release the dataset and code via the GitHub: \url{https://github.com/Exploration-Lab/PoseStich-SLT}.
\end{itemize}

\vspace{-2mm}
\section{Related Work} \label{sec:related}
\vspace{-2mm}

SLT has gained increasing interest in recent years \cite{phoenix_dataset, info13050220, De_Coster_2021_AT4SSL, chen2022two-stream,joshi-etal-2023-isltranslate}. Most prior works \cite{yin2020attention, yin-read-2020-better, moryossef2021data} rely heavily on intermediate gloss annotations to enable supervised training of translation systems. The gloss-based approaches assume a two-stage pipeline: first mapping sign videos to glosses, and then translating glosses into spoken language. While effective, they require extensive manual annotation and may strip away linguistic richness.
Early work in end-to-end sign language translation, such as STMC-Transformer \cite{yin-read-2020-better}, extends the STMC architecture originally developed for gloss-level recognition \cite{9354538}. However, a central challenge across these models is the token length mismatch between input sign video frames and output textual tokens \cite{lin2023glossfreeendtoendsignlanguage, camgoz2020sign, Alvarez2022SignLT}. This issue complicates sequence alignment, especially in encoder-decoder frameworks, where long visual input sequences must be compressed to shorter textual outputs.
To mitigate this mismatch, several approaches have been proposed. Some rely on learning intermediate gloss-like representations through frame clustering or joint training objectives \cite{camgoz2020sign, chen2022simple}, while others \cite{ahn2024slowfast, yin-read-2020-better} modify the architecture by introducing conceptual anchors or applying CTC loss for better temporal alignment \cite{10.1145/1143844.1143891}. GloFE \cite{lin2023glossfreeendtoendsignlanguage} is a recent gloss-free system that addresses these alignment challenges using pose-based inputs. It introduces weak intermediate representations derived from spoken-language tokens (termed ``conceptual anchors'') to supervise visual feature learning. GloFE is closely related to our target setting, as it also uses keypoint-based inputs and bypasses gloss labels. Our work differs in both approach and emphasis. Instead of relying on architectural modifications or auxiliary representations, we propose a linguistically driven pretraining strategy that leverages publicly available word-level sign language datasets. Using grammar templates from BLiMP, we synthesize millions of sentence-level training examples by stitching pose sequences from CISLR (ISL) \cite{joshi-etal-2022-cislr} and WLASL (ASL) \cite{li2020word}. Despite using a vanilla encoder-decoder Transformer, our model achieves performance comparable to or better than specialized models like GloFE, demonstrating the effectiveness and scalability of our approach.

\vspace{-2mm}
\section{Methodology} \label{sec:methods}
\vspace{-2mm}

\begin{table*}[t]
    \centering
    \resizebox{\linewidth}{!}{
    \tiny
    \setlength{\tabcolsep}{5pt}
    \begin{tabular}{l l l cccc cccc}
        \toprule
        & \textbf{Method} & \textbf{Source} & \multicolumn{4}{c}{\textbf{DEV}} & \multicolumn{4}{c}{\textbf{TEST}} \\
        \cmidrule(lr){4-7} \cmidrule(lr){8-11}
        & & & BLEU-1 & BLEU-2 & BLEU-3 & BLEU-4 & BLEU-1 & BLEU-2 & BLEU-3 & BLEU-4 \\
        \midrule
        \multirow{7}{*}{\rotatebox{90}{How2Sign (ASL)}} 
        & Alvarez & -- & 17.73 & 7.94 & 4.13 & 2.24 & 17.40 & 7.69 & 3.97 & 2.21 \\
        & GloFE & -- & 14.92 & 7.13 & 3.90 & 2.27 & 14.86 & 6.99 & 3.64 & 1.97 \\
        \cmidrule(lr){2-11}
        & w/o Pose Stitched & -- & 22.19 & 9.71 & 5.43 & 3.37 & 18.16 & 7.96 & 4.33 & 2.62 \\
        
        & \multirow{2}{*}{Pose Stitched (RWO)} 
        & BPCC & 26.25 & 13.00 & 7.50 & 4.62 & 26.70 & \textbf{12.90} & 7.23 & 4.32 \\
        & & BLIMP & 26.22 & 13.06 & 7.81 & \textbf{5.04} & 25.17 & 12.26 & 7.15 & 4.53 \\
        & \multirow{2}{*}{Pose Stitched (SWO)} 
        & BPCC & \textbf{27.35} & \textbf{13.56} & 7.85 & 4.95 & \textbf{26.74} & 12.86 & \textbf{7.26} & 4.44 \\
        & & BLIMP & 26.89 & 13.36 & \textbf{7.92} & \textbf{5.04} & 25.98 & 12.55 & 7.25 & \textbf{4.56} \\
        \cmidrule(lr){4-11}
        & Best (Ours) & -- & \textbf{27.35 (+9.62)}  & \textbf{13.56 (+5.62)}  & \textbf{7.92 (+3.79)}  & \textbf{5.04 (+2.77)}  & \textbf{26.74 (+9.34)}  & \textbf{12.90 (+5.21)}  & \textbf{7.26 (+3.29)}  & \textbf{4.56 (+2.35)}  \\
        
        \midrule
        \multirow{7}{*}{\rotatebox{90}{iSign (ISL)}} 
        & GloFE & -- & 8.92 & 2.85 & 1.18 & 0.61 & 9.20 & 2.83 & 1.12 & 0.55 \\
        
        \cmidrule(lr){2-11}
        & w/o Pose Stitched & -- & 12.85 & 2.50 & 1.04 & 0.58 & 12.81 & 2.66 & 1.14 & 0.64 \\
        & \multirow{2}{*}{Pose Stitched (RWO)}
        & BPCC & 14.75 & 6.11 & 3.60 & 2.45 & 14.89 & 6.25 & 3.67 & 2.51 \\
        & & BLIMP & 13.89 & 4.86 & 2.57 & 1.62 & 13.39 & 4.63 & 2.43 & 1.48 \\
        & \multirow{2}{*}{Pose Stitched (SWO)}
        & BPCC & \textbf{17.31} & \textbf{8.09} & \textbf{5.02} & \textbf{3.54} & \textbf{17.67} & \textbf{8.20} & \textbf{5.00} & \textbf{3.43} \\
        & & BLIMP & 16.42 & 7.51 & 4.62 & 3.23 & 16.44 & 7.40 & 4.49 & 3.09 \\
        \cmidrule(lr){4-11}
            & Best (Ours) & -- & \textbf{17.31 (+8.39)}  & \textbf{8.09\ (+5.24)}  & \textbf{5.02 (+3.84)}  & \textbf{3.54 (+2.93)}   & \textbf{17.67 (+8.47)}  & \textbf{8.20 (+5.37)}  & \textbf{5.00 (+3.88)} & \textbf{3.43 (+2.88)} \\
        \bottomrule
    \end{tabular}}
    \caption{BLEU score results on the How2Sign and iSign datasets comparing different pretraining strategies (random word order vs. same word order), along with baseline results (GloFE and Alvarez). Corresponding ROUGE scores are provided in App. Table \ref{tab:rouge_results}. The numbers in brackets show the absolute improvements from the baseline.}
    \vspace{-6mm}
    \label{tab:merged_results}
\end{table*}

\noindent\textbf{Problem Setup:}
We consider the task of translating sign language videos, represented as 2D pose sequences, into spoken language text. Formally, given a sequence of frame-wise pose vectors $\mathcal{X} = \{x_1, x_2, \dots, x_T\}$ extracted from a sign language video, where each $x_t \in \mathbb{R}^{152}$ is a concatenation of keypoints from the face, hands, and upper body, the goal is to generate a corresponding English sentence $\mathcal{Y} = \{y_1, y_2, \dots, y_N\}$. 
In contrast to prior work that depends on intermediate gloss annotations or signer-identifiable video, we directly learn from pose-level inputs to preserve privacy and avoid costly manual labeling. 

\noindent\textbf{Data Generation via Linguistic Templates and Large Text Corpora:}
To overcome the dataset scarcity, we construct synthetic training datasets by leveraging grammatical templates covering a wide range of linguistic phenomena. Our key insight is that by aligning linguistic templates with word-level sign pose data, we can synthesize large numbers of sentence-pose pairs suitable for model pretraining. We take inspiration from the BLiMP benchmark \cite{warstadt-etal-2020-blimp-benchmark}, which captures a broad set of linguistic phenomena (67 paradigms across 12 categories). Using these templates, we generate millions of grammatically diverse English sentences. However, to synthesize paired sign sequences, we require corresponding pose representations for each word in the targeted sign language. To this end, we leverage two publicly available word-level sign language datasets: WLASL \cite{li2020word} for American Sign Language (ASL) and CISLR \cite{joshi-etal-2022-cislr} for Indian Sign Language (ISL). We identify the overlapping vocabulary between BLiMP and each of these datasets, 508 words for WLASL and 504 for CISLR, and construct two synthetic datasets: \textbf{1) 
BLiMP-ASL:} \textit{\textbf{2.8 million}} English sentences using the shared vocabulary between WLASL and BLiMP. \textbf{2) BLiMP-ISL:} \textbf{\textit{22 million}} English sentences using the shared vocabulary between CISLR and BLiMP.  For example, consider the Adjunct Island paradigm from BLiMP Templates \cite{warstadt-etal-2020-blimp-benchmark}: 

\noindent \texttt{Wh[] Aux\_mat[] Subj[] V\_mat[] Adv[] V\_emb[] Obj[] }

\noindent By filling this template with shared vocabulary, we can obtain sentences like:
\texttt{``What did John read before filing the book?''} In this example, the words like 
\texttt{``what,'' ``John,'' ``read,'' ``filing,'' and ``book,''} are drawn from the overlapping vocabulary
of BLiMP–CISLR or BLiMP–WLASL and the syntactic structure is governed by the BLiMP template.
\noindent Our sentence generation process combines linguistically motivated templates with sign language vocabulary. Since sign languages are less studied in the literature and we do not yet have a comprehensive understanding of sign language grammar, particularly in low-resource settings such as ISL, we ground the sentence generation process in English grammar, under the assumption that there exist underlying correlations between spoken and sign language grammars. In doing so, we ensure that the resulting sentences are both grammatically well-formed and systematically associated with specific linguistic phenomena. (see App. \ref{app:dataset_detailsA1} for more details).

\noindent Though the BLiMP-based generation provides strong grammatical diversity, its vocabulary coverage is limited. To increase linguistic and lexical coverage, we complement this with data from the BPCC corpus \cite{BPCC}, a large collection of \textbf{\textit{230 million}} English bitext pairs. From this, we select sentences that have a word match of over 90\% with the WLASL or CISLR vocabulary, thereby ensuring that they can be fully synthesized using the available sign poses. This yields two additional datasets: BPCC-ASL and BPCC-ISL. Further post-processing (e.g., sentence length matching  (see App. \ref{app:dataset_detailsA3} for more details)., anonymization via token replacement  (see App. \ref{app:dataset_detailsA4} for more details), and filtering infrequent words) ensures compatibility with downstream training objectives. Further details and dataset statistics are presented in App. \ref{app:dataset_details}. \\
\noindent\textbf{Pose Stitching:} 
To synthesize sentence-level sign language sequences, we stitch together word-level pose sequences. For each sentence, we retrieve individual word videos from WLASL or CISLR and extract 2D keypoints using the Mediapipe library \cite{MediaPipe}, covering facial expressions, hand configurations, and upper body motion. We select 76 keypoints (forming 152-dimensional vectors) that most effectively capture sign-relevant articulation, inspired by prior work \cite{lin2023glossfreeendtoendsignlanguage}. Low-confidence keypoints are interpolated to maintain consistency, and all sequences are normalized to reduce inter-signer variance. \\
\noindent The stitched pose sequences are constructed by temporally aligning and concatenating word-level segments into a fluent stream. For smooth transitions between signs and to avoid abrupt motion boundaries, we apply boundary-aware temporal smoothing \cite{bulas1993temporal} 
The resulting pose sequences emulate coherent signing while retaining compositional structure. \\
\noindent A key design choice is the word order used during pose stitching. Unlike spoken languages, most sign languages have a distinct and often non-linear grammatical structure that differs significantly from English. However, due to the lack of accessible linguistic resources, reliable syntactic parsers, or annotated corpora for ASL or ISL grammar, we do not attempt to reconstruct native sign language word order in our synthetic data. Instead, we adopt English word order as a proxy, which simplifies generation and leverages the available textual infrastructure. Further details regarding framerate matching and Pose processing are discussed in the App. \ref{app:pose_stichingB1} , \ref{app:pose_stichingB2} . \\
\noindent To explore how sensitive the model is to this assumption, we construct two variants of our synthetic datasets: \textbf{1) \textit{Same Word Order (SWO):}} Poses are stitched in the same order as the English sentence, preserving syntactic structure and compositional cues. \textbf{2) \textit{Random Word Order (RWO):}} Poses are stitched after randomly permuting the word order, injecting syntactic noise, and encouraging the model to learn flexible and robust representations. These two variants allow us to investigate the trade-off between syntactic alignment and generalization, especially in low-resource or cross-lingual sign language translation settings. \\
\noindent To effectively leverage the synthetic datasets and ensure a smooth transition to real-world data, we also employ a linear annealing strategy during training (see App. \ref{app:architecturalC2} for more details). Initially, the model is trained exclusively on synthetic pose-sentence pairs. As training progresses, we gradually increase the probability of sampling from the real sentence-aligned datasets, iSign for ISL and How2Sign for ASL, up to a threshold of 85\% at 60,000 training steps (also see App. \ref{fig:isign-complementary-sampling}). After this point, training continues predominantly on real data (sign language pose-sequences from iSign and How2Sign datasets for ISL and ASL, respectively). 
This staged training approach helps the model benefit from both the diversity and scale of synthetic data and the realism of target domain data. We found that this strategy works better than the traditional disjoint pretraining and fine-tuning phases. Unlike traditional pretraining-fine-tuning pipelines with disjoint phases, our curriculum is blended and progressive, allowing for continual adaptation and preventing catastrophic forgetting. The full architecture (Transformer-based encoder-decoder), tokenization strategy, and hyperparameters are detailed in the App. \ref{app:section-2}.

\vspace{-2mm}
\section{Results and Analysis} \label{sec:results}
\vspace{-2mm}

To evaluate the effectiveness of our synthetic pose-based pretraining strategy, we train a  Transformer encoder-decoder model \cite{NIPS2017_3f5ee243} on two benchmark datasets: How2Sign (ASL) and iSign (ISL).

\noindent\textbf{Translation Performance}:
We report BLEU-4 \cite{papineni-etal-2002-bleu} and ROUGE-L \cite{lin-2004-rouge} scores for both datasets, following evaluation protocols from prior work \cite{lin2023glossfreeendtoendsignlanguage} to ensure fair comparison. 
As shown in Table \ref{tab:merged_results}, our approach outperforms the previous state-of-the-art, GloFE, by a substantial margin: On How2Sign, BLEU-4 improves from $2.27 \rightarrow 5.04$ on the dev set and $1.97 \rightarrow 4.56$ on the test set.
On iSign, BLEU-4 improves from $0.61 \rightarrow 3.54$ on the dev set and $0.55 \rightarrow 3.43$ on the test set.
We also observe consistent improvements in ROUGE scores across both datasets (see App. Table \ref{tab:rouge_results}), further validating the quality of generated translations. We also report the SacreBLEU scores of the best model (see App. Table \ref{tab:sacre_results}).  
While the scores remain modest, the consistent gains demonstrate the promise of our approach for improving SLT in low-resource sign languages.

\noindent\textbf{Ablation/Impact of Synthetic Data:}
To isolate the impact of synthetic pose-stitched pretraining, we conduct a baseline experiment using the same architecture and hyperparameters, but \textit{\textbf{without any synthetic data}}. As shown in App. Fig. \ref{fig:Before-after-PT}, this variant performs significantly worse, highlighting the critical role of the proposed pretraining pipeline.

\noindent\textbf{Generalization to Unseen Sentences:}
To test generalization, we evaluate our pretrained models on new pose-stitched sentences not seen during training. Surprisingly, the model achieves BLEU-4 scores of 97 and 47 on two synthetic evaluation sets (App. Fig. \ref{fig:bleu_scores_pretraining}), indicating that the model learns to robustly handle sentence generation within the restricted vocabulary domain.

\noindent\textbf{Similarity to Target Domain:}
To further understand domain alignment, we compute semantic similarity scores between sentences in the synthetic and target datasets using SBERT (all-MiniLM-L6-v2) \cite{huggingfaceSentencetransformersallMiniLML6v2Hugging}. App. Table \ref{tab:most-least-similar} shows that higher-similarity examples correlate with better translation quality, reinforcing the effectiveness of our dataset construction approach.

\noindent\textbf{Qualitative Results:}
App. Table~\ref{tab:ref-pred-table-How2sign} presents qualitative examples of predicted translations alongside ground truth references. Our model consistently captures the main semantic content and shows better alignment than GloFE. However, occasional issues such as phrase repetition or hallucinations remain, likely due to synthetic data noise. 
Additional results are provided in the App. \ref{app: Additional Results} . The effect of adding the pose-stitched dataset is discussed in App. \ref{app: Additional ResultsD1}, while the effect of distribution shift is analyzed in App. \ref{app: Additional ResultsD2}. Qualitative analyses and the impact of random versus same word order in the pretraining dataset are presented in the App. \ref{app: Additional ResultsD3} and App. \ref{app: Additional ResultsD4}.


\noindent\textbf{Grammatical Notion of Generated Sign Language Sentences}: Understanding and modeling sign language grammar is a difficult task. Sign language relies on visual-spatial structures, non-manual markers, role shifts, and the use of signing space, making its grammar highly complex \cite{sinha2017indian,joshi-etal-2024-isign}. Standardized grammatical resources for low-resource languages like ISL are limited. Similar challenges are noted for ASL, where resources such as \cite{sehyr2021asl} and recent work \cite{tavella-etal-2022-wlasl-lex} only begin to capture aspects of its grammar. These complexities highlight that directly modeling sign language grammar is non-trivial. This motivates us to ground synthetic data generation in English grammatical templates while acknowledging the gap between spoken and signed language structures.

\vspace{-2mm}
\section{Conclusion and Future Directions} \label{sec:conclusion}
\vspace{-2mm}


In this work, we introduce a novel training paradigm for sign language translation by leveraging existing word-level pose datasets to synthesize sentence-level training data. Our approach constructs pose-stitched sentence sequences using linguistically grounded templates, enabling large-scale pretraining without requiring expensive gloss annotations or raw video footage. Through extensive experiments on two benchmark datasets, How2Sign (ASL) and iSign (ISL), we demonstrate that this strategy significantly improves translation performance, even when using a standard Transformer-based encoder-decoder model. \\ 
\noindent This work opens new avenues for scaling sign language translation using linguistic structure-based data synthesis. Future efforts may explore expanding the vocabulary coverage of word-level pose datasets, integrating grammatical features from sign languages, and applying this strategy across more diverse sign language variants to build more inclusive and generalizable SLT systems.

\vspace{-2mm}
\section*{Limitations} \label{sec:limitations}
\vspace{-2mm}
Despite the improvements demonstrated by our training strategy, showing a significant boost in performance when compared to existing SOTA methods, there are several limitations that remain open:
\noindent\textbf{Vocabulary Coverage:}
Our method relies on the intersection of BLiMP vocabulary and available word-level sign datasets (WLASL and CISLR), which cover only 2K–4.5K words. This restricts the expressiveness of synthetic sentences and limits generalization. Extending word-level datasets to include broader and more diverse vocabularies remains critical, but was beyond the scope of this work due to a lack of publicly available resources.

\noindent\textbf{Absence of Sign Language Grammar:}
We rely on English word order for generating pose-stitched sequences, as grammatical annotations for sign languages are limited and not standardized. Although this introduces potential mismatches, modeling sign-specific syntax would require extensive linguistic resources and annotation efforts, which are currently lacking for most sign languages.

\noindent\textbf{Architectural Simplicity:}
We adopt a standard transformer model to isolate the effect of our training strategy. While this ensures clarity in evaluation, it may underutilize recent advances in sign-specific architectures. 
This choice was made intentionally to enable scalable pretraining without relying on proprietary datasets or gloss annotations. 
Moreover, we could not fully replicate prior methods (e.g., GloFE) due to unavailable details, limiting direct comparisons. Furthermore, our framework is modular and adaptable, designed to make the most of available resources while remaining extensible as new sign language datasets become available, which further adds to the advantage of using simple architectures.

\noindent\textbf{Generality Across Languages:}
Although our approach is applicable beyond ASL and ISL, broader adoption depends on the availability of word-level datasets in other sign languages. Resource creation in this space remains a foundational challenge.


\vspace{-2mm}
\section*{Ethical Considerations}
\vspace{-3mm}


Our work focuses on the task of sign language translation, with an emphasis on both American Sign Language (ASL) and Indian Sign Language (ISL). The goal is to use technology to enhance the daily lives of the deaf and hard-of-hearing communities in both regions. While we have made improvements over previous models, the proposed system still lacks the capability to function as a fully realized interpreter in real-life scenarios. We use extracted key points as the input for the model, ensuring minimal to no concerns regarding personal privacy.
\vspace{-2mm}
\section*{Acknowledgments}
\vspace{-2mm}
We would like to thank the anonymous reviewers and the meta-reviewer for their insightful comments and suggestions. 
This research work was partially supported by the Research-I Foundation of the Department of CSE at IIT Kanpur. 

\bibliography{references}

\clearpage
\newpage

\appendix

\section*{Appendix}

\appendix


\titlecontents{section}[18pt]{\vspace{0.05em}}{\contentslabel{1.5em}}{}
{\titlerule*[0.5pc]{.}\contentspage} 


\titlecontents{table}[0pt]{\vspace{0.05em}}{\contentslabel{1em}}{}
{\titlerule*[0.5pc]{.}\contentspage} 

\startcontents[appendix] 
\section*{Table of Contents} 
\printcontents[appendix]{section}{0}{\setcounter{tocdepth}{4}} 

\startlist[appendix]{lot} 
\section*{List of Tables} 
\printlist[appendix]{lot}{}{\setcounter{tocdepth}{1}} 

\startlist[appendix]{lof} 
\section*{List of Figures} 
\printlist[appendix]{lof}{}{\setcounter{tocdepth}{1}} 

\newpage

\begin{figure*}[h]
  \centering
  \includegraphics[width=0.95\linewidth]{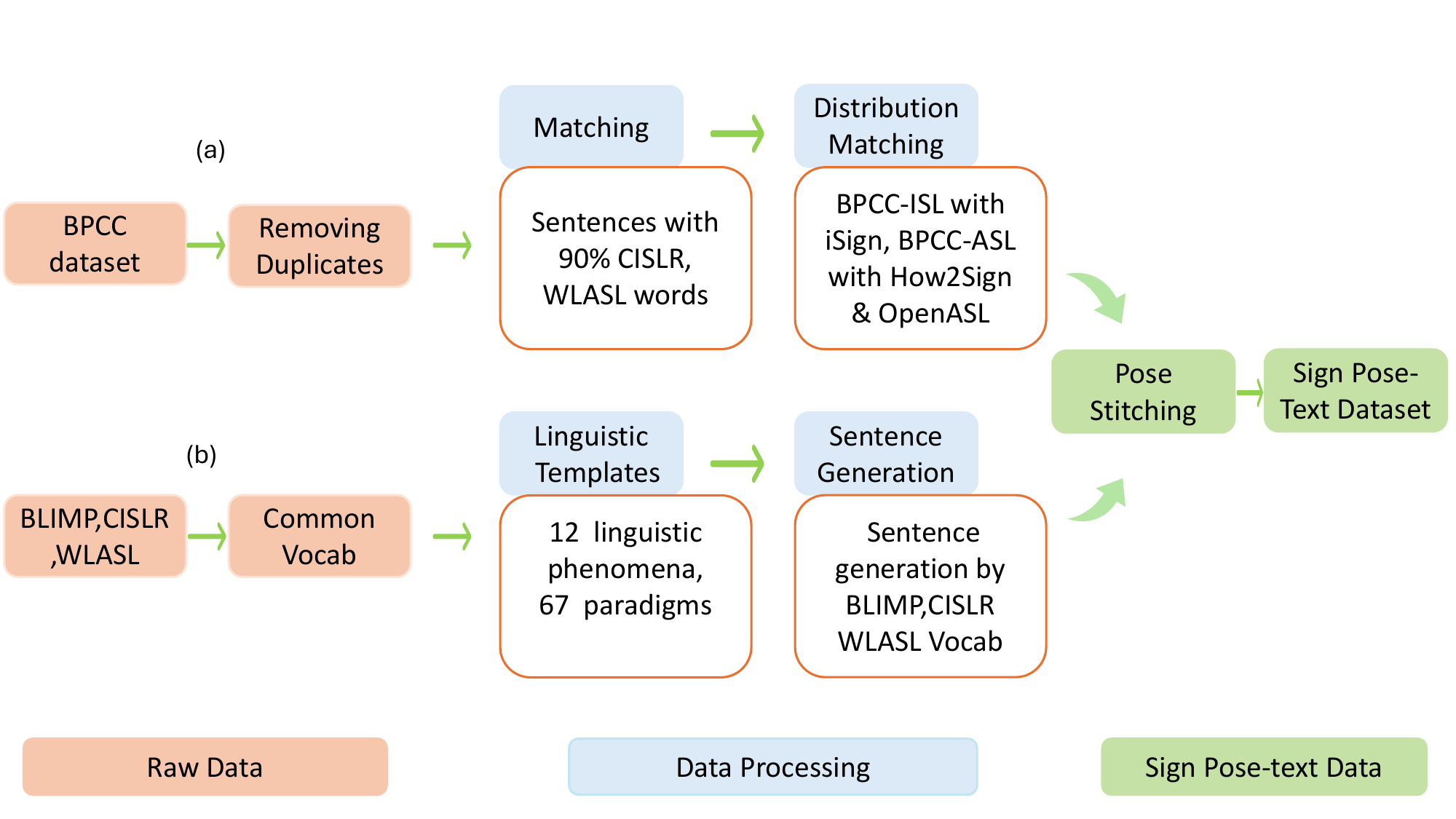}
  \caption[Data Creation Pipeline using different sources]{(a) From the filtered raw BPCC Corpus, sentences are matched with CISLR and WLASL vocabulary, and the sentence length distribution is aligned with iSign and How2Sign. Two datasets, \BPCCISL\ and \BPCCASL\, are created after pose stitching (b). Using templates from BLIMP \cite{warstadt-etal-2020-blimp-benchmark} and the common vocabulary of BLIMP, CISLR, and WLASL, two additional datasets are created: \BLIMPISL\ and \BLIMPASL\ after pose stitching.}
  \label{fig:data-creation-pipeline}
\end{figure*}

\section{Pose Stitched Dataset Details} \label{app:dataset_details}

For the generated set of sentences, we found that the linguistic-template-based sentences cover a small set of words available in the corresponding sign language datasets (WLASL and CISLR). To explore further, we create another set using sentences from a translation dataset, BPCC \cite{BPCC}. This helps increase the coverage of words, essentially improving the overlap with the target sign language datasets (How2Sign and iSign). Overall, we create four datasets using the proposed pose strategy: two using the linguistic templates from BLIMP \cite{warstadt-etal-2020-blimp-benchmark}, and two using the large corpus of translation text, BPCC \cite{BPCC}.

Further, we perform additional preprocessing to match the generated sentence distribution with the target distribution to make the training more effective. In this section, we discuss the details of the entire dataset formation pipeline. An overview of the pipeline is represented in Fig. \ref{fig:data-creation-pipeline}. In general, we have two target sign languages (ASL from How2Sign and ISL from iSign). We create the additional sentences using pose stitching to match the target distribution (both in terms of vocabulary as well as sequence length). We discuss each of the steps in the pipeline below:


\subsection{Dataset created using linguistic Templates}
\label{app:dataset_detailsA1}
We utilize the common vocabulary of BLIMP, CISLR, and WLASL for sentence generation. We generated the sentences by linguistic templates proposed in Benchmark of Linguistic Minimal Pairs (BLiMP) \cite{warstadt-etal-2020-blimp-benchmark}, which covers 67 linguistic paradigms with 12 phenomena. In total, we generated 22M sentences using the common vocabulary of BLIMP and CISLR. Similarly, using the common vocabulary of BLIMP and WLASL, we generated around 2.8M sentences. The dataset generated by linguistic templates using the common vocabulary of CISLR and BLIMP is referred to as \BLIMPISL. The dataset generated by linguistic templates using the common vocabulary of WLASL and BLIMP is referred to as \BLIMPASL. A sample of a few generated sentences is shown in Table \ref{tab:linguistic-phenomena} with their respective linguistic phenomenon.

\begin{table*}[t]
\centering
\resizebox{\textwidth}{!}
{
\begin{tabular}{llcc}
\toprule
Phenomenon & N & \BLIMPISL & \BLIMPASL \\
\midrule
ANAPHOR AGR. & 2 & \textit{Some people will hide them-
selves.} & \textit{ Many children will fire themselves.} \\
ARG. STRUCTURE & 9 & \textit{Some boy should clean.}  & \textit{Some woman will stretch that jacket.}
 \\
BINDING & 7 & \textit{some actress might think that she
can wear that blouse.} & \textit{It's herself that a girl can admire.}\\
CONTROL/RAISING & 5 & \textit{that boy will predict it to be bad
that all children wear this skirt.} & \textit{that girl will find it to be bad that every grandmother can wear a coat.}
\\
DET.-NOUN AGR. & 8 & \textit{These children drink that beer.}  & \textit{that boy should wear this big coat.}
 \\
ELLIPSIS & 2 & \textit{some boy will wear one big
blouse and that girl will wear
three.} & \textit{some boy will wear one big coat and a girl will wear two.}
\\
FILLER-GAP & 7 &\textit{ some boy will see some coat that people wear.} & \textit{every boy can know that some boy should wear this coat.}
 \\
IRREGULAR FORMS & 2 & - & - \\
ISLAND EFFECTS & 8 & \textit{Who should that father and some
teacher visit? } & \textit{who can boy's building this bank aid.}
 \\
NPI LICENSING & 7 & \textit{Even people will often appear.} &\textit{Even that cousin should also admire this pie.}
 \\
QUANTIFIERS & 4 & \textit{that nephew can insult at least
five children.} & \textit{that brother should purchase at least six socks.}
 \\
SUBJECT-VERB AGR. & 6 & \textit{The people hide these people. }& - \\
\bottomrule
\end{tabular}
}
\caption[Linguistic Phenomenon]{Twelve Different Phenomenon from \cite{warstadt-etal-2020-blimp-benchmark} and sentences from \BLIMPISL\ and \BLIMPASL\ dataset }
\label{tab:linguistic-phenomena}
\end{table*}

\subsection{Dataset Created using BPCC Corpus }
\label{app:dataset_detailsA2}
 The BPCC corpus \cite{BPCC} is a large dataset with 230 million bitext pairs. We matched the words in the sentences with the CISLR and WLASL vocabulary and selected the sentences with more than $90\%$ word match. After selecting the sentences, we performed sentence merging(we merged small sentences) to match the length distribution of the sentences with the How2Sign and iSign datasets.

\subsection{Sentence Length Matching}
\label{app:dataset_detailsA3}
\begin{figure*}[t]
  \centering
  \begin{minipage}[b]{0.48\textwidth}
    \centering
    \includegraphics[width=\textwidth]{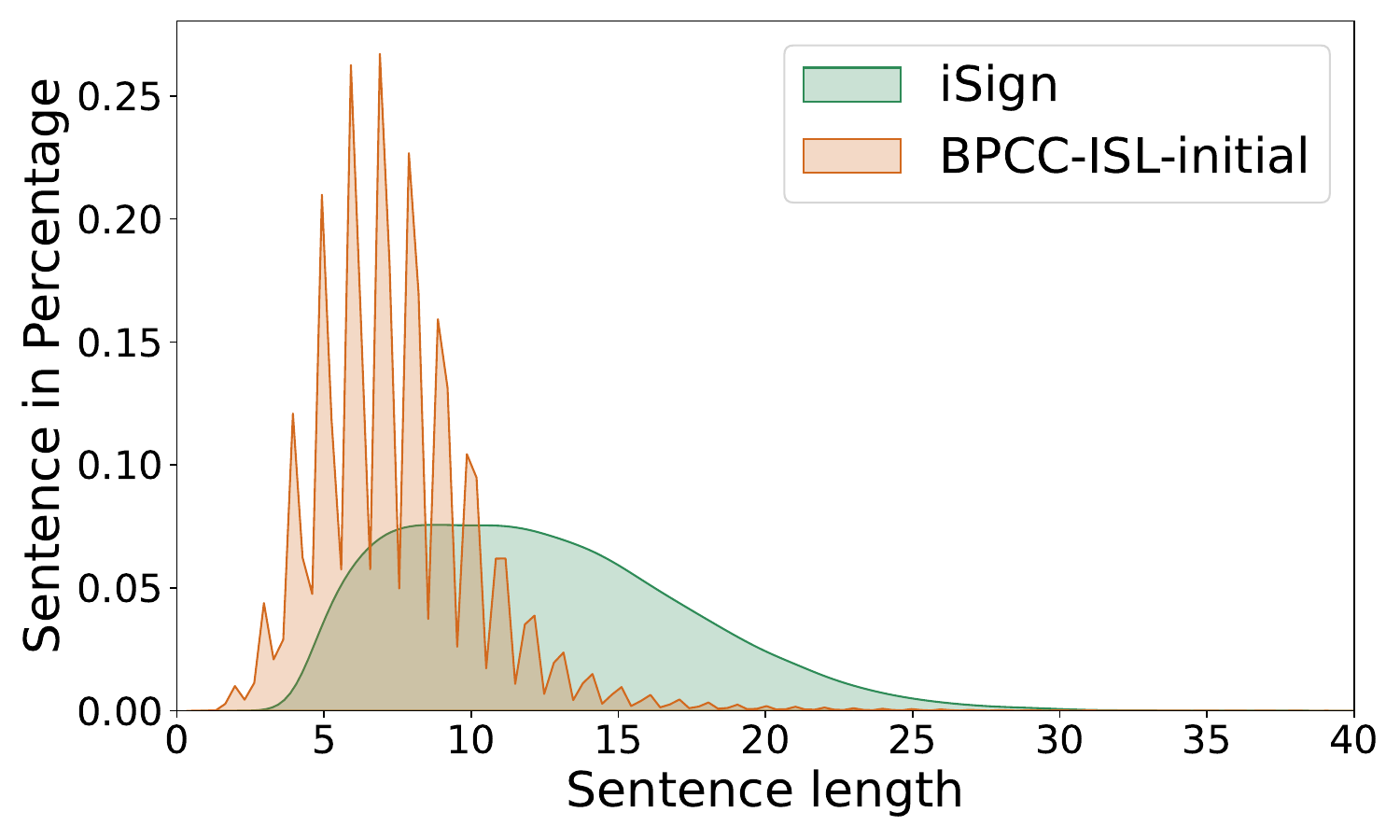}
    \caption[Sentence length distribution ISL-Initial]{Sentence length distribution between iSign train and \BPCCISL\ dataset before merging sentences}
    \label{fig:isign-Synthetic_CISLR-before}
  \end{minipage}
  \hfill
  \begin{minipage}[b]{0.48\textwidth}
    \centering
    \includegraphics[width=\textwidth]{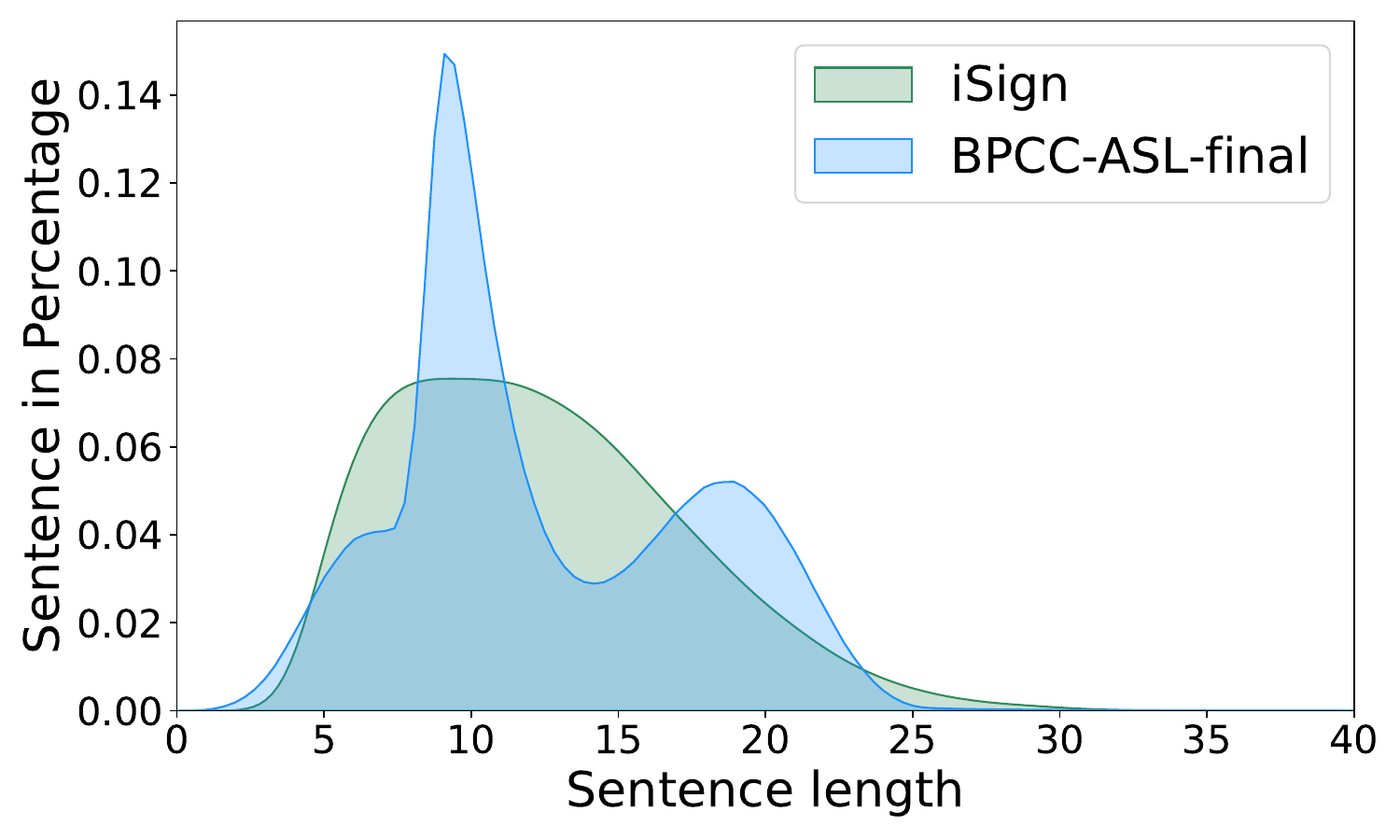}
    \caption[Sentence length distribution ISL-final]{Sentence length distribution between iSign train and \BPCCISL\ dataset after merging sentences}
    \label{fig:isign-Synthetic_CISLR-after}
  \end{minipage}

  \vspace{0.5cm}
  
  \begin{minipage}[b]{0.48\textwidth}
    \centering
    \includegraphics[width=\textwidth]{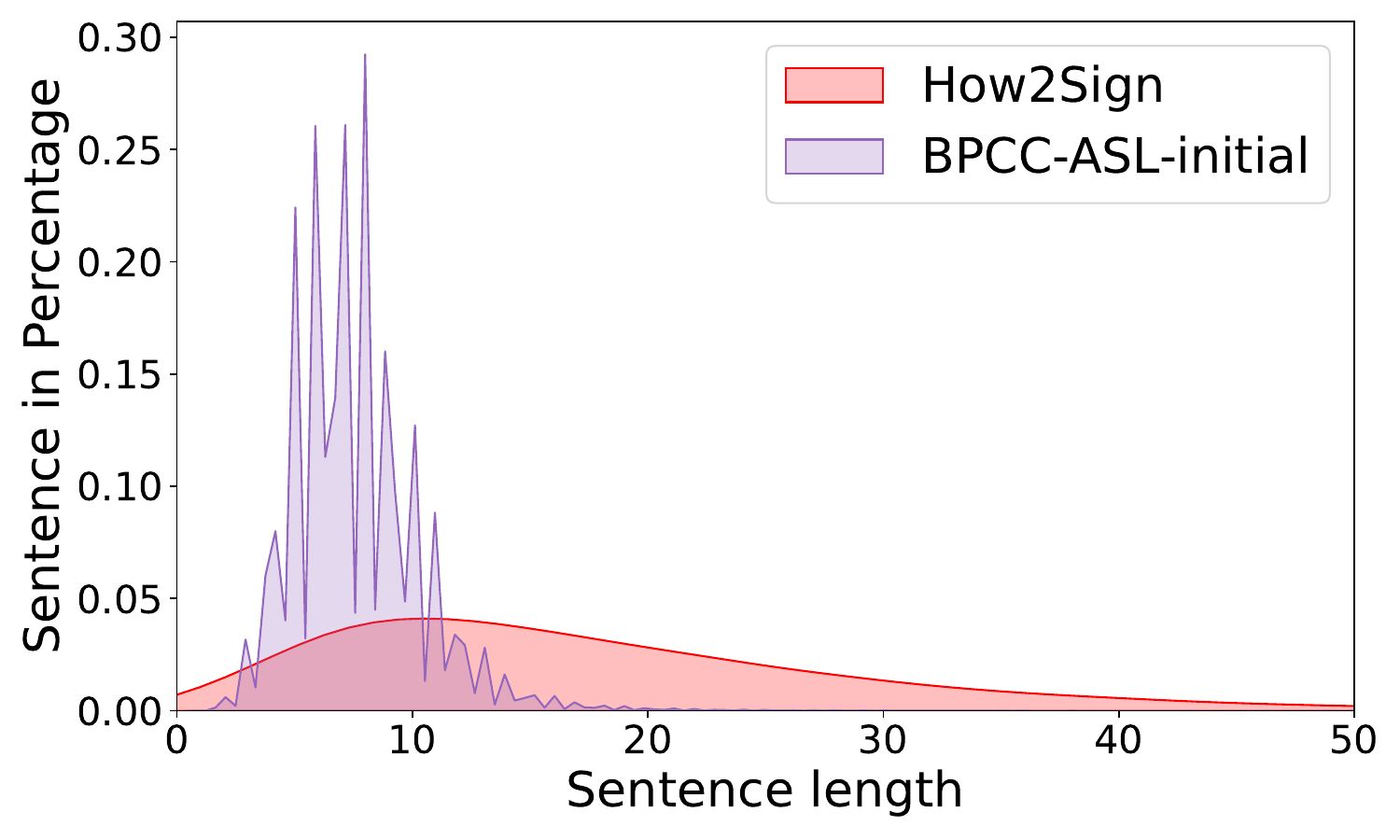}
    \caption[Sentence length distribution ASL-Initial]{Sentence length distribution between How2Sign and \BPCCASL\  dataset before merging sentences}
    \label{fig:How2sign-Synthetic_WLASL-before}
  \end{minipage}
  \hfill
  \begin{minipage}[b]{0.48\textwidth}
    \centering
    \includegraphics[width=\textwidth]{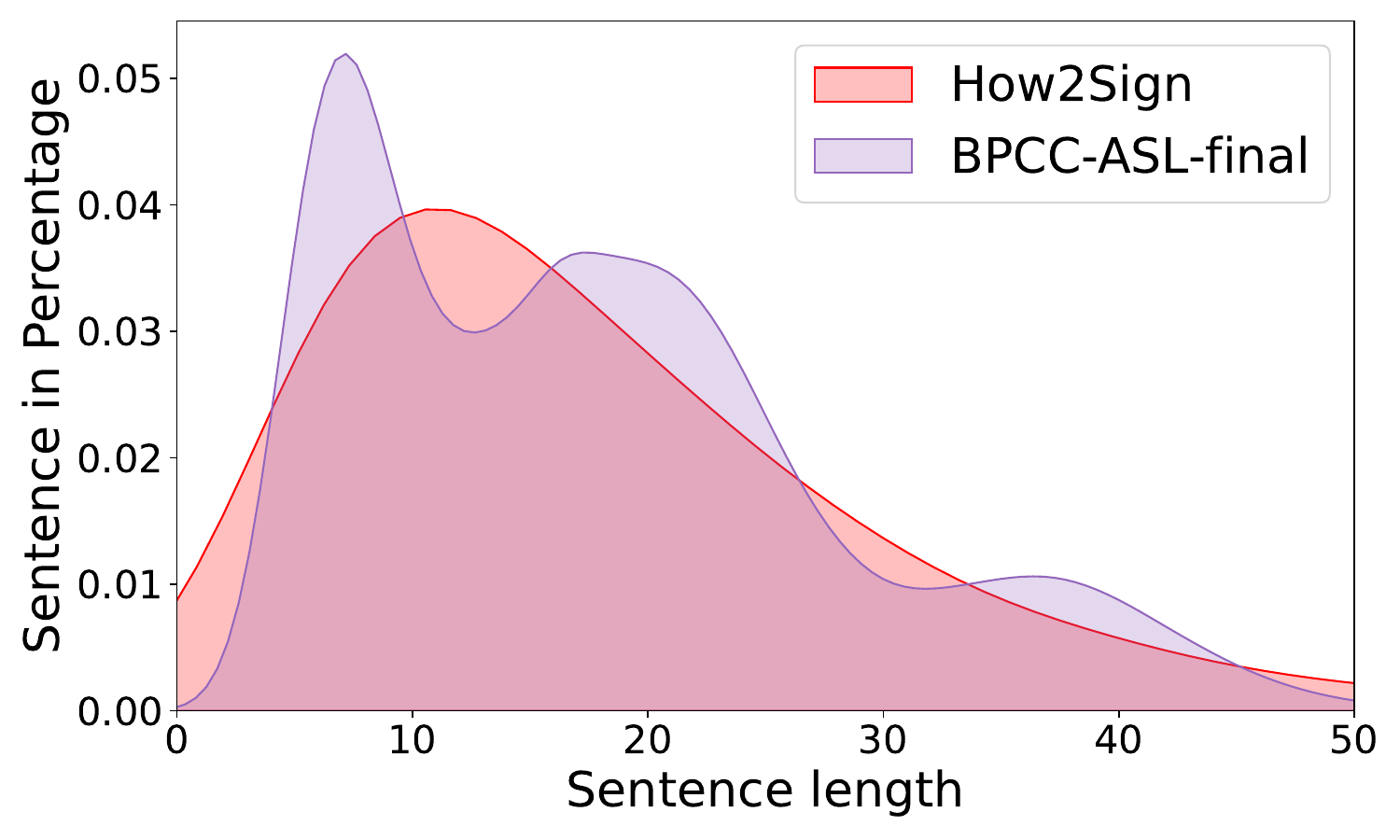}
    \caption[Sentence length distribution ISL-Final]{Sentence length distribution How2Sign and \BPCCASL\ after merging sentences}
    \label{fig:How2sign-Synthetic_WLASL-after}
  \end{minipage}
    \hfill
 \begin{minipage}[b]{0.48\textwidth}
  \centering
  
  \includegraphics[width=\linewidth]{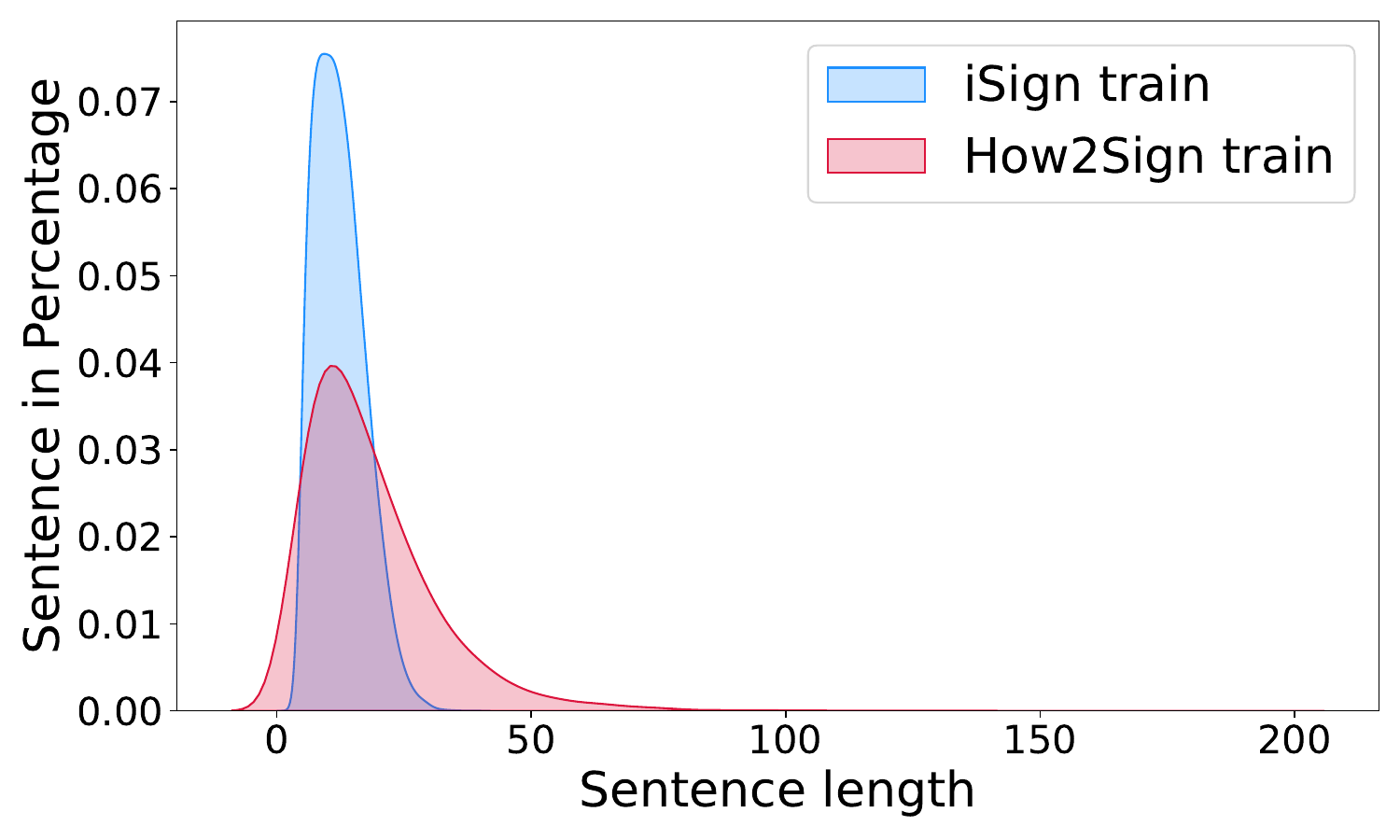}
  \caption{Sentence length distribution between How2Sign and iSign dataset}
  \label{fig:isign-How2Sign}
\end{minipage}
  \label{fig:combined-sentence-distributions}
\end{figure*}

To match the sentence length distribution of the iSign dataset, we took 90\% of the sentences from the dataset created using BPCC and CISLR with a length of less than 8, and merged three sentences into one. This merging and remaining 10\% of the sentences resulted in 1.6M sentences, which we will refer to as the \BPCCISL\ dataset. The sentence length distribution between the \BPCCISL\ and iSign datasets before merging sentences is shown in Fig. \ref{fig:isign-Synthetic_CISLR-before}, and after merging sentences is shown in Fig. \ref{fig:isign-Synthetic_CISLR-after}.
Similarly, different combinations of sentence merging on the dataset created by matching BPCC and WLASL resulted in 1.1 million sentences. We will refer to this dataset as \BPCCASL. The sentence length distribution between \BPCCASL\ and the How2Sign dataset before merging sentences is shown in Fig. \ref{fig:How2sign-Synthetic_WLASL-before}, and after merging is shown in Fig. \ref{fig:How2sign-Synthetic_WLASL-after}.

\subsection{Post-processing} 
\label{app:dataset_detailsA4}
We further post-process the data by replacing the name of Person with <PERSON> and words with frequency less than 3 with <UNKNOWN> in the dataset (\BPCCISL, \BPCCASL)  and from the train set of How2Sign and the train set of iSign.

\subsection{Summary of Created Datasets}
\label{app:dataset_detailsA5}
\begin{figure*}[h]
  \centering
  \begin{minipage}[b]{0.38\textwidth}
    \centering
    \includegraphics[width=\textwidth]{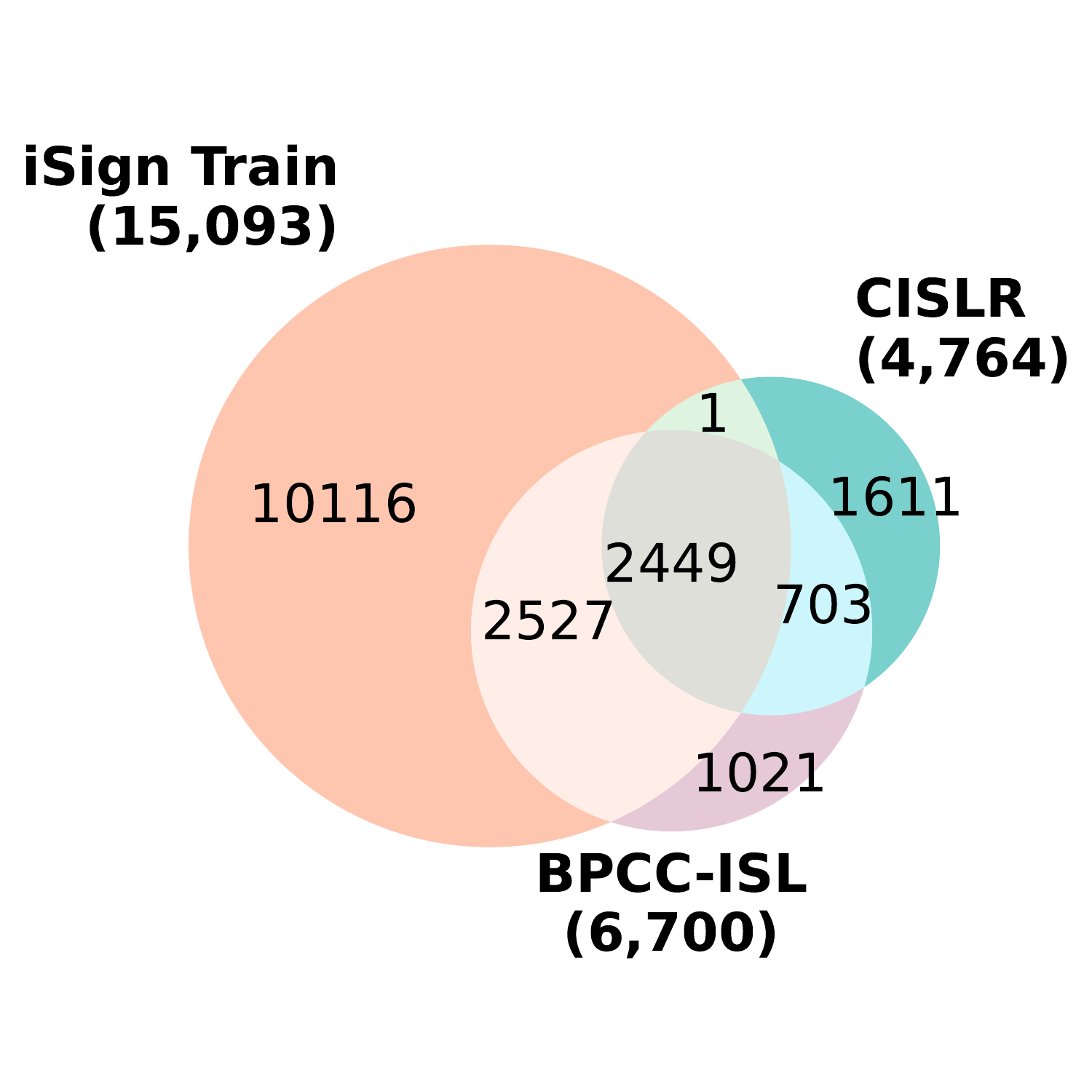}
    \caption{Vocabulary Distribution between CISLR, iSign train, \BPCCISL }
    \label{fig:vocab-isign-cislr-synthetic-cislr}
  \end{minipage}
  \hfill
  \begin{minipage}[b]{0.38\textwidth}
    \centering
    \includegraphics[width=\textwidth]{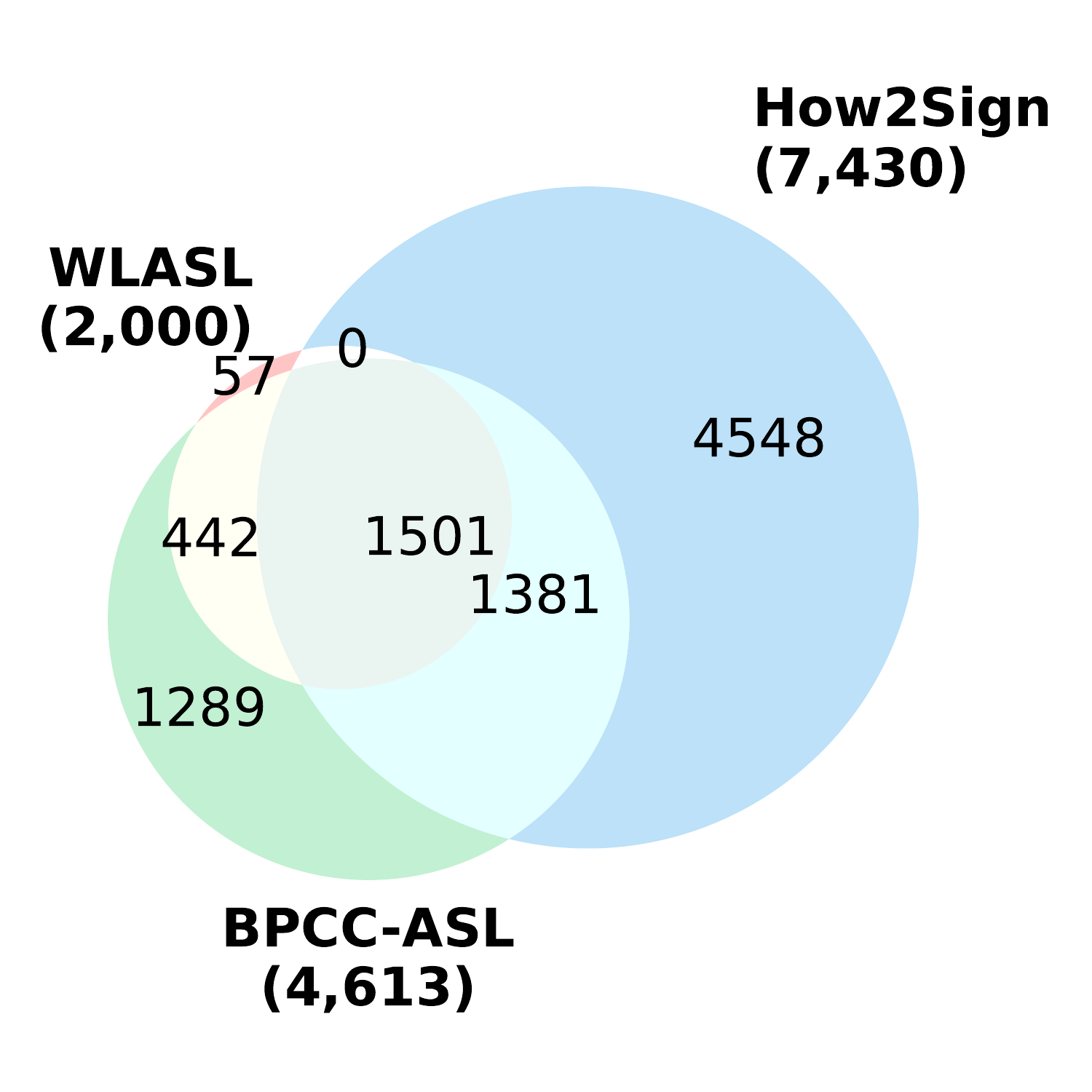}
    \caption{Vocabulary Distribution between WLASL, How2Sign, \BPCCASL}
    \label{fig:vocab-wlasl-how2sign-synthetic-wlasl}
  \end{minipage}

  \vspace{0.1cm}
  
  \begin{minipage}[b]{0.38\textwidth}
    \centering
    \includegraphics[width=\textwidth]{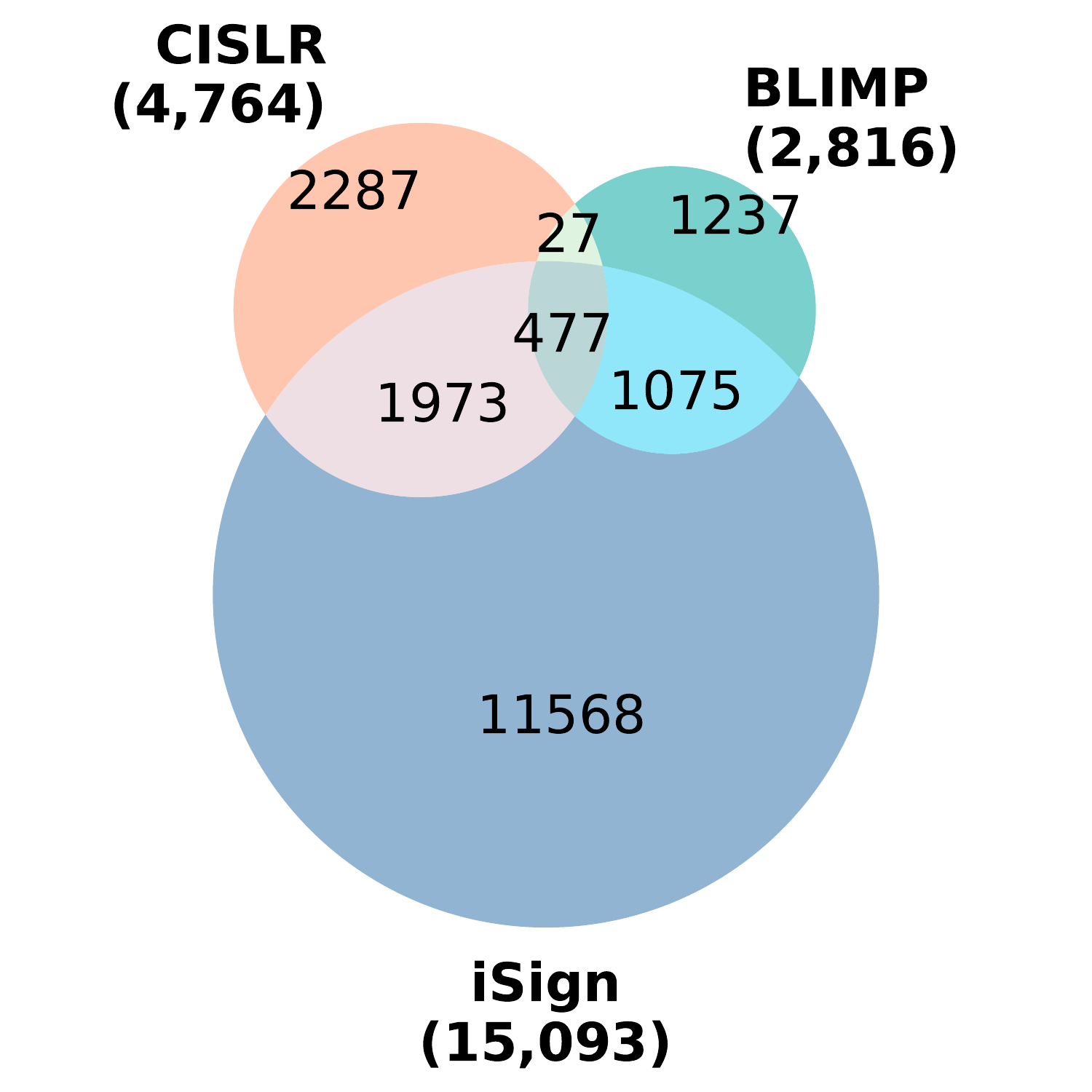}
    \caption{Vocabulary Distribution between BLIMP, CISLR, iSign train}
    \label{fig:vocab-cislr-blimp-isign}
  \end{minipage}
  \hfill
  \begin{minipage}[b]{0.38\textwidth}
    \centering
    \includegraphics[width=\textwidth]{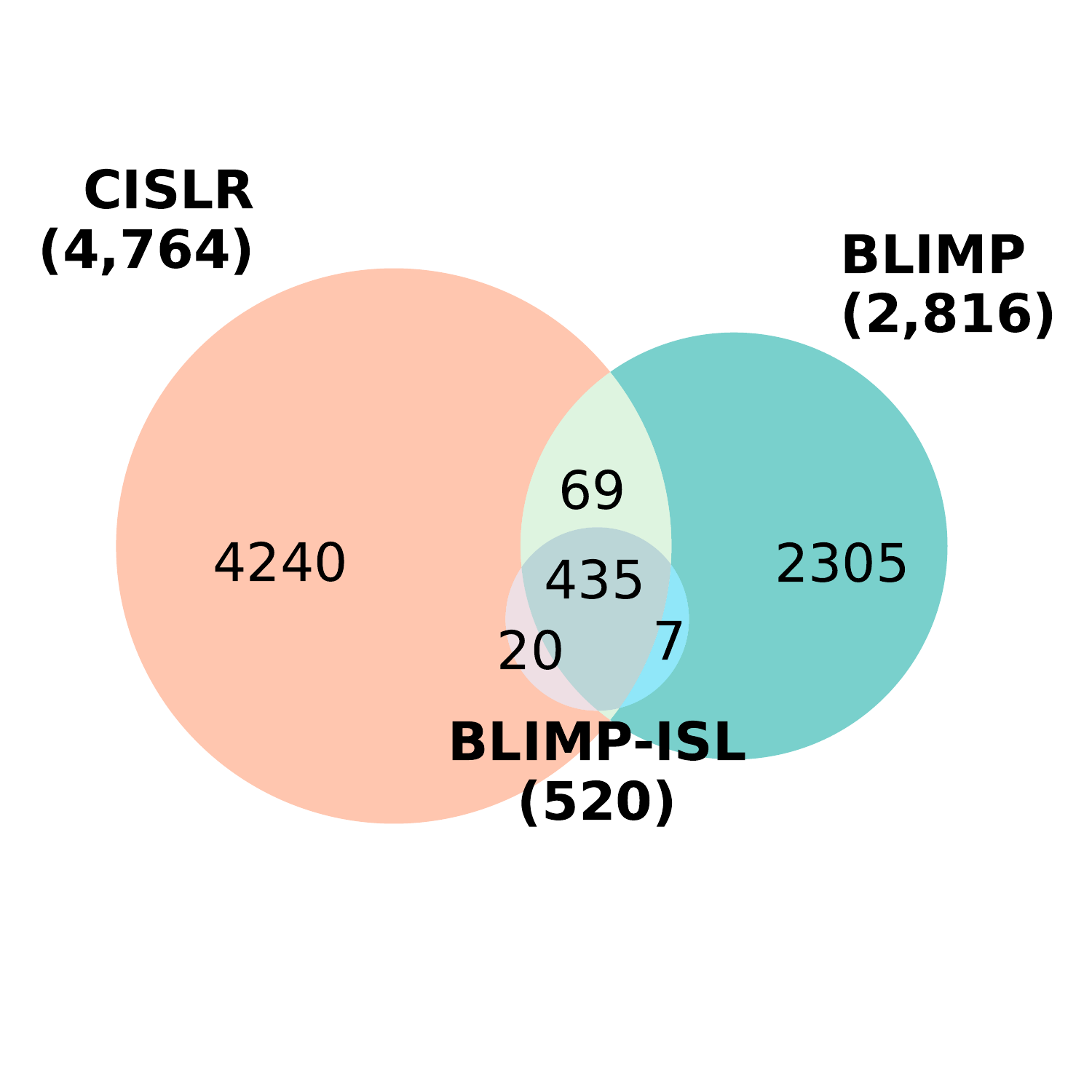}
    \caption{Vocabulary Distribution between CISLR, BLIMP, \BLIMPISL}
    \label{fig:vocab-cislr-blimp-synthetic_cislr_blimp}
  \end{minipage}

  \vspace{0.1cm}
  
  \begin{minipage}[b]{0.38\textwidth}
    \centering
    \includegraphics[width=\textwidth]{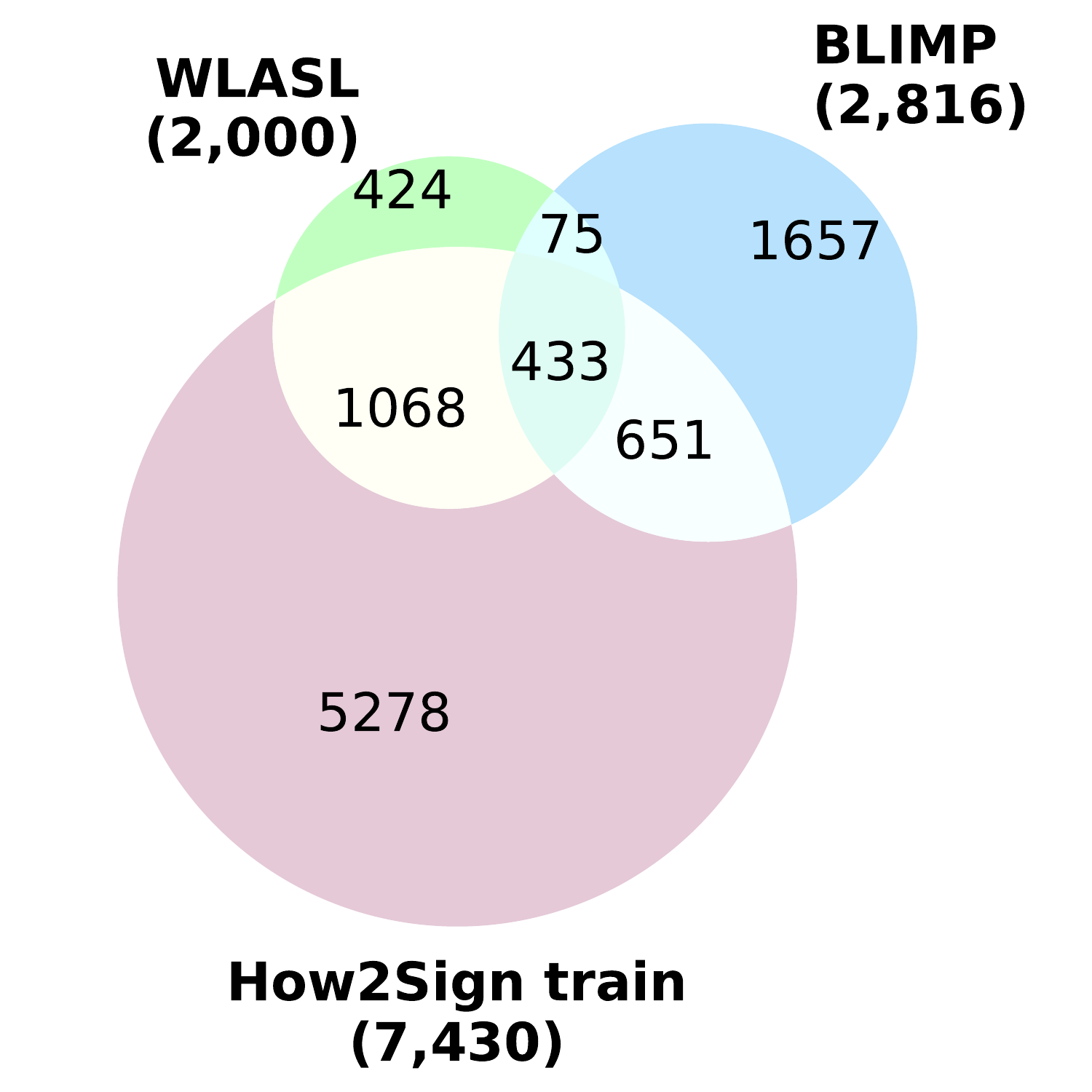}
    \caption{Vocabulary Distribution between WLASL, BLIMP, How2Sign}
    \label{fig:vocab-wlasl-blimp-how2sign}
  \end{minipage}
  \hfill
  \begin{minipage}[b]{0.38\textwidth}
    \centering
    \includegraphics[width=\textwidth]{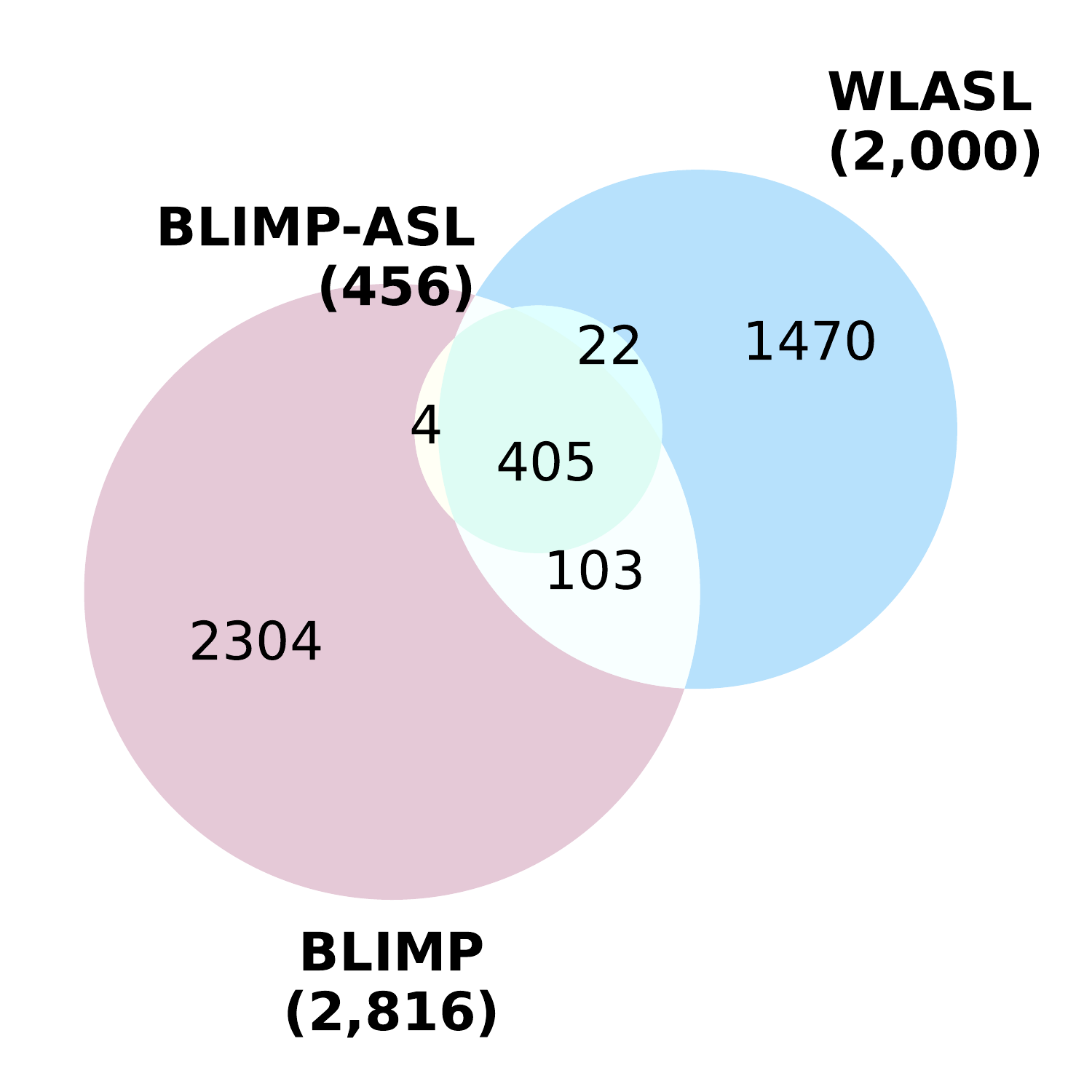}
    \caption{Vocabulary Distribution between \BLIMPASL\, WLASL, BLIMP}
    \label{fig:vocab-Synthetic_WLASL_Blimp-WLASL-Blimp}
  \end{minipage}
  \label{fig:combined-vocab-distributions}
\end{figure*}

In total, we have two vocabulary data sets, CISLR and WLASL, four pose stitched datasets (\BLIMPISL, \BLIMPASL, \BPCCISL, \BPCCASL\ ),  How2Sign (train, test, val), iSign (train, test, val). The number of sentences in each dataset is shown in Table \ref{tab:dataset_sizes}. Vocabulary count in iSign Train, CISLR, How2Sign, WLASL, \BPCCISL, BPCC-ASL presented in Table \ref{tab:unique_words_overlap} venn diagram of common vocab is shown in Fig.\ref{fig:vocab-isign-cislr-synthetic-cislr}, Fig.\ref{fig:vocab-wlasl-how2sign-synthetic-wlasl}. Vocabulary count and common vocab between CISLR, BLIMP, iSign, \BLIMPISL\ is presented in Table \ref{tab:unique_words_CISLR_BLIMP_iSign-synthetic-cislr-blimp} and a Venn diagram of common vocab is shown in Fig. \ref{fig:vocab-cislr-blimp-isign} and Fig. \ref{fig:vocab-cislr-blimp-synthetic_cislr_blimp}. Vocabulary count and common vocab between  WLASL, BLIMP, How2Sign, \BPCCASL\  presented in Table \ref{tab:unique_words_WLASL_BLIPM_how2sign_SYnthetic_CISLR_BLIMP} and a Venn diagram of common vocab is shown in Fig. \ref{fig:vocab-wlasl-blimp-how2sign} and Fig. \ref{fig:vocab-Synthetic_WLASL_Blimp-WLASL-Blimp}.

\begin{table}[t]
    \centering
    \small
    {
    \begin{tabular}{cc}
        \toprule
        \textbf{Dataset Name} & \textbf{Number of Sentences} \\
        \midrule
        iSign Train & 99,923 \\
        iSign Test & 6,069 \\
        iSign val & 5,653 \\
        How2Sign Train & 31,092 \\
        How2Sign Test & 2,349 \\
        How2Sign Val & 1,739 \\
        \BPCCISL& 1,640,469 \\
        \BPCCASL& 1,173,536 \\
        \BLIMPISL& 22,219,407 \\
        \BLIMPASL & 2,880,008 \\
        \bottomrule
    \end{tabular}
    }
    \caption{Number of Sentences in Each Dataset.}
    \label{tab:dataset_sizes}
\end{table}

    \begin{table}[t]
    \centering
    \small
    {
    \begin{tabular}{lc}
        \toprule
        Dataset Combination & Unique Words \\
        \midrule
        iSign Train & 15,093 \\
        CISLR & 4,764 \\
        \BLIMPISL & 520 \\
        BLIMP & 2,816 \\
        iSign Train $\cap$ CISLR & 2,450 \\
        iSign Train $\cap$ \BLIMPISL & 457 \\
        iSign Train $\cap$ BLIMP & 1,552 \\
        CISLR $\cap$ \BLIMPISL & 455 \\
        CISLR $\cap$ BLIMP & 504 \\
        \BLIMPISL $\cap$ BLIMP & 442 \\
        iSign Train $\cap$ CISLR $\cap$ \BLIMPISL & 439 \\
        iSign Train $\cap$ CISLR $\cap$ BLIMP & 477 \\
        iSign Train $\cap$ \BLIMPISL $\cap$ BLIMP & 426 \\
        CISLR $\cap$ \BLIMPISL $\cap$ BLIMP & 435 \\
        All datasets & 419 \\
        \bottomrule
    \end{tabular}
    }
    \caption[Unique word counts ISL]{Unique word counts in CISLR , BLIMP, iSign-Train, \BLIMPISL }
    \label{tab:unique_words_CISLR_BLIMP_iSign-synthetic-cislr-blimp}
\end{table}
\begin{table}[t]
    \centering
    \small
    \resizebox{\columnwidth}{!}
    {
    \begin{tabular}{lc}
        \toprule
        Dataset Combination & Unique Words \\
        \midrule
        \BLIMPASL & 456 \\
        WLASL & 2,000 \\
        BLIMP & 2,816 \\
        How2Sign Train & 7,430 \\
        \BLIMPASL $\cap$ WLASL & 427 \\
        \BLIMPASL $\cap$ BLIMP & 409 \\
        \BLIMPASL $\cap$ How2Sign Train & 380 \\
        WLASL $\cap$ BLIMP & 508 \\
        WLASL $\cap$ How2Sign Train & 1,501 \\
        BLIMP $\cap$ How2Sign Train & 1,084 \\
        \BLIMPASL $\cap$ WLASL $\cap$ BLIMP & 405 \\
        \BLIMPASL $\cap$ WLASL $\cap$ How2Sign Train & 366 \\
        \BLIMPASL $\cap$ BLIMP $\cap$ How2Sign Train & 348 \\
        WLASL $\cap$ BLIMP $\cap$ How2Sign Train & 433 \\
        WLASL $\cap$ BLIMP $\cap$ How2Sign Train $\cap$ \BLIMPASL & 344 \\
        \bottomrule
    \end{tabular}
    }
    \caption[Unique word counts ASL]{Unique word counts in WLASL , BLIMP, How2Sign-Train, \BLIMPASL }
    \label{tab:unique_words_WLASL_BLIPM_how2sign_SYnthetic_CISLR_BLIMP}
\end{table}

\begin{table}[h]
    \centering
    \resizebox{\columnwidth}{!}{
    \begin{tabular}{lc}
        \toprule
        Dataset Combination & Unique Words \\
        \midrule
        iSign Train & 15,093 \\
        CISLR & 4,764 \\
        \BPCCISL & 6,700 \\
        iSign Train $\cap$ CISLR & 2,450 \\
        iSign Train $\cap$ \BPCCISL & 4,976 \\
        CISLR $\cap$ \BPCCISL & 3,152 \\
        CISLR $\cap$ isign Train $\cap$ \BPCCISL & 2,449 \\
        WLASL & 2,000 \\
        How2Sign Train & 7,430 \\
        \BPCCASL & 4,613 \\
        WLASL $\cap$ How2Sign Train & 1,501 \\
        WLASL $\cap$ \BPCCASL & 1,943 \\
        How2Sign Train $\cap$ \BPCCASL & 2,882 \\
        WLASL $\cap$ How2Sign Train $\cap$ \BPCCASL  & 1,504 \\
        \bottomrule
    \end{tabular}
    }
    \caption[Unique word counts ISL, ASL]{Unique word counts for isign Train, CISLR, \BPCCISL, How2Sign Train, WLASL, \BPCCASL\ dataset and their overlaps}
    \label{tab:unique_words_overlap}
\end{table}

\begin{figure*}[t]
  \centering
  \begin{minipage}[b]{0.46\textwidth}
    \centering
    \includegraphics[width=\textwidth]{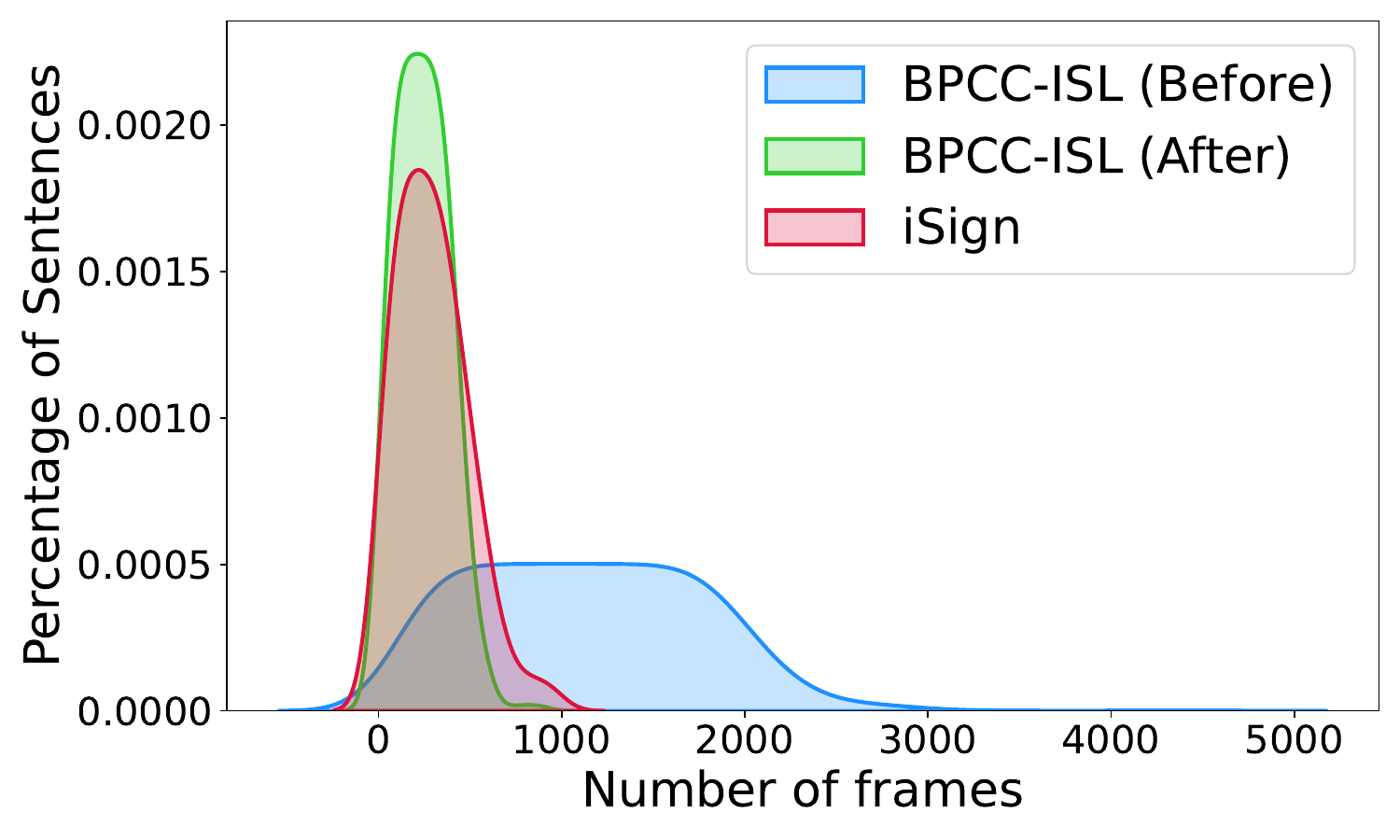}
    \caption{Frame Distribution between \BPCCISL\ , iSign dataset}
    \label{fig:Frame-distribution-Synthetic-CISLR-isign}
  \end{minipage}
  \hspace{0.8cm}
    \begin{minipage}[b]{0.46\textwidth}
    \centering
    \includegraphics[width=\textwidth]{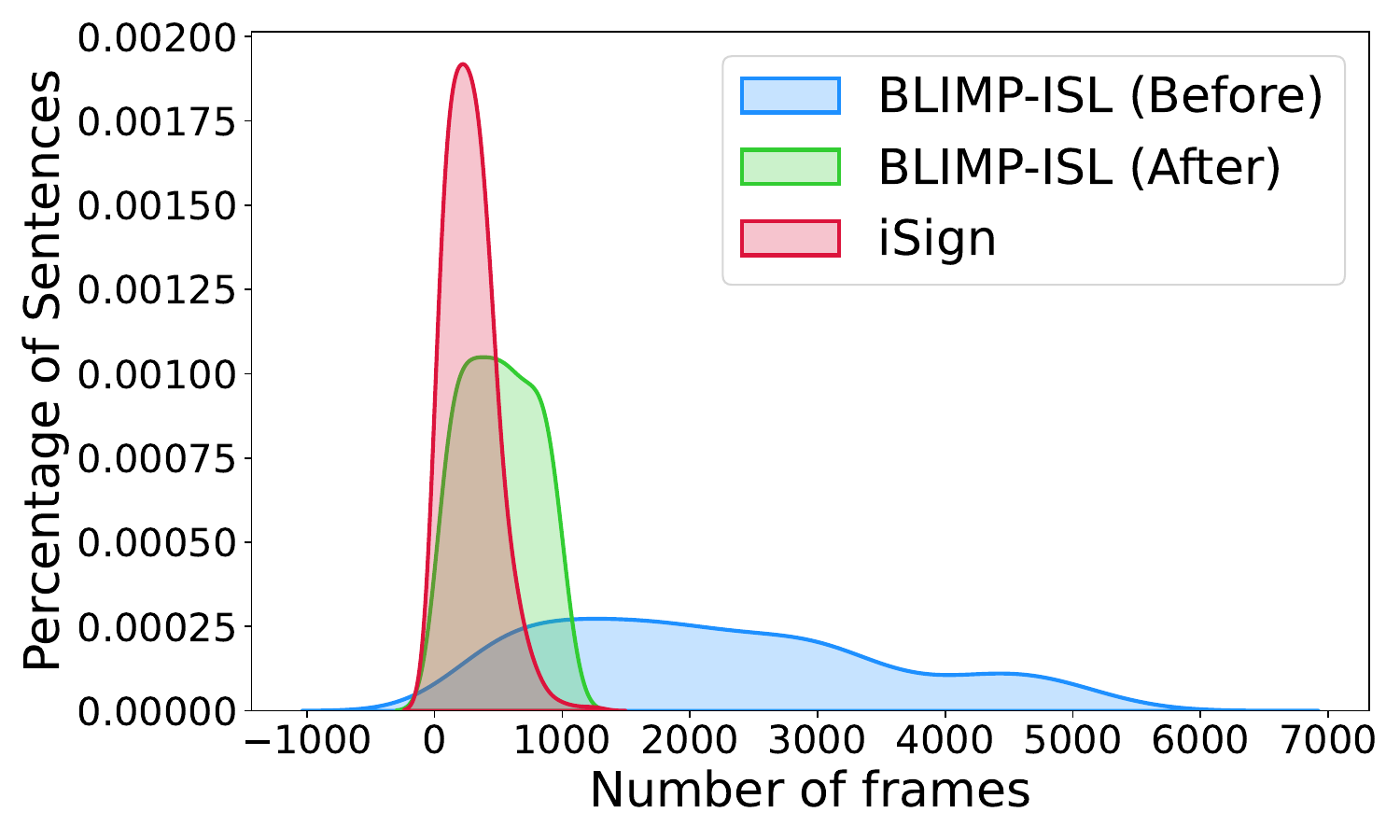}
    \caption{Frame Distribution between \BLIMPISL\ , iSign dataset}
    \label{fig:Frame-Distribution-Synthetic-CISLR-Blimp-iSign}
  \end{minipage}
  \begin{minipage}[b]{0.46\textwidth}
    \centering
    \includegraphics[width=\textwidth]{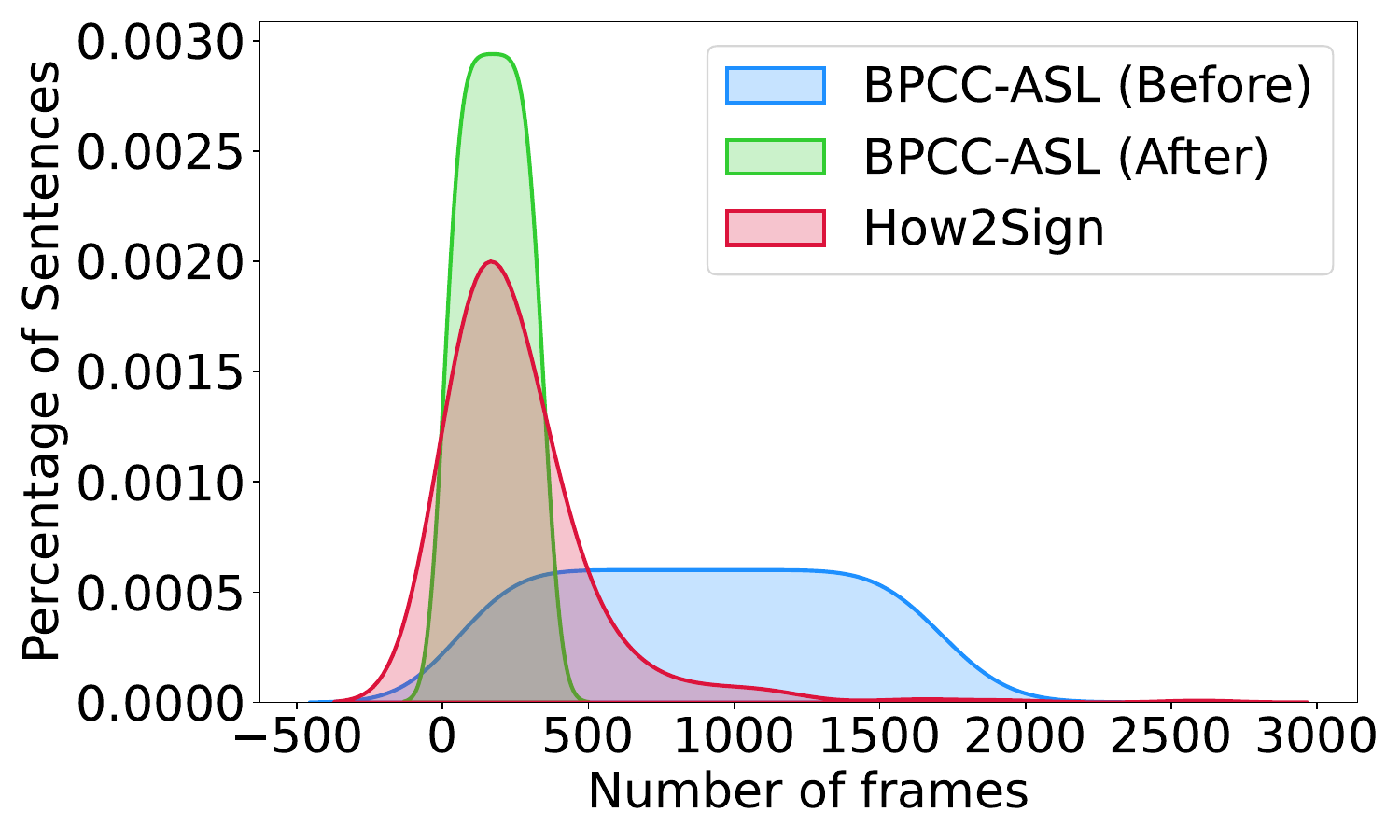}
    \caption{Frame Distribution between \BPCCASL\ , How2Sign dataset}
    \label{fig:Frame-distribution-Synthetic_WLASL-How2Sign}
  \end{minipage}
  \hspace{0.8cm}
  \begin{minipage}[b]{0.46\textwidth}
    \centering
    \includegraphics[width=\textwidth]{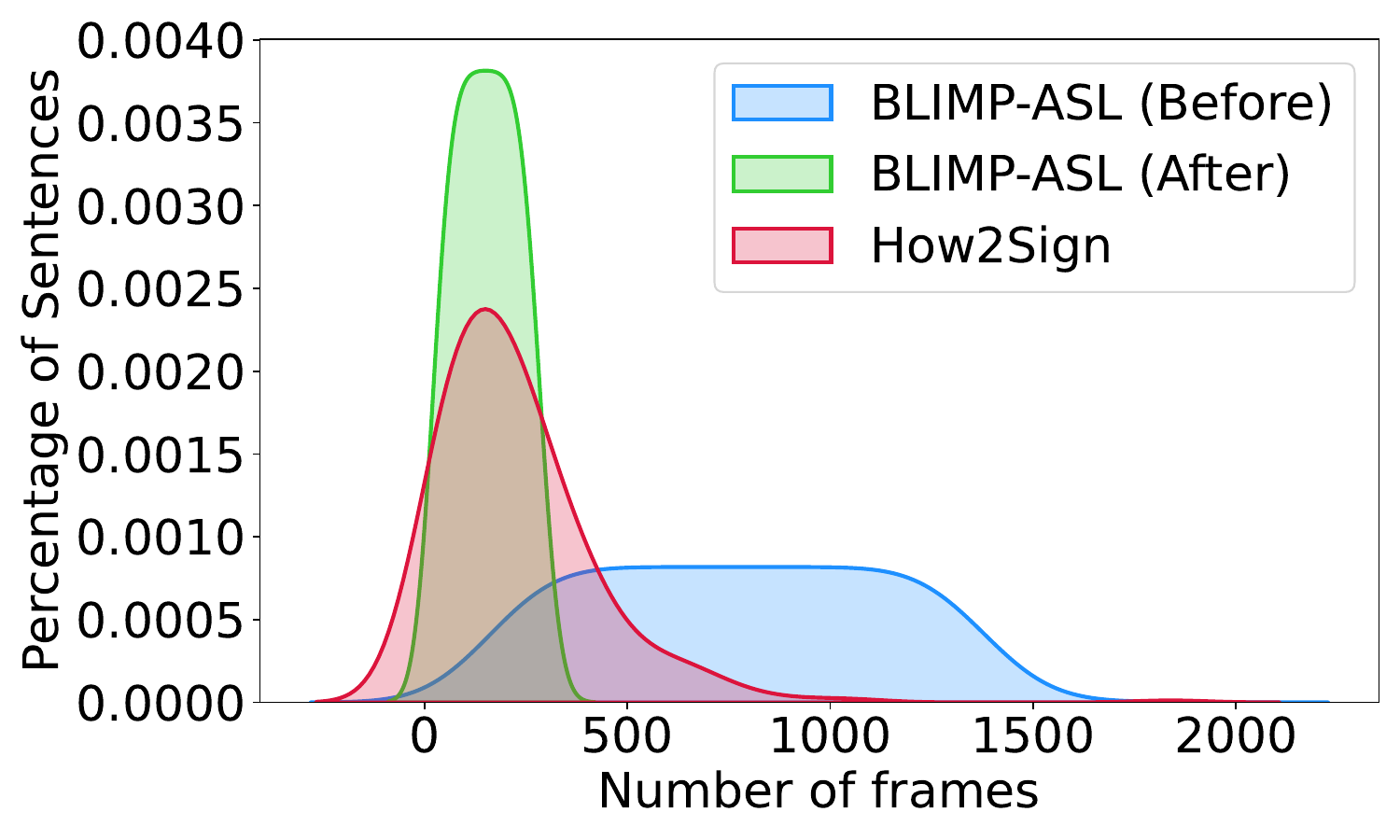}
    \caption{Frame Distribution between \BLIMPASL\ , How2Sign dataset}
    \label{fig:Frame-Distribution-Synthetic-WLASL-Blimp-How2Sign}
  \end{minipage}

  \label{fig:combined-frame-distributions}

  \label{fig:combined-frame-distributions}
\end{figure*}

\section{Processing Stitched Poses}
\label{app:pose_stiching}
\subsection{Framerate Matching} 
\label{app:pose_stichingB1}
To match the frame rate of the two sources(i.e, for example \BPCCISL\ and iSign), we made the mean of the two sources the same and sampled every xth frame. Where x  denotes the sampling rate described in the \ref{app:section-2}. Frame distribution between \BPCCISL\ and iSign dataset before and after framerate matching is present in Fig. \ref{fig:Frame-distribution-Synthetic-CISLR-isign}. Similarly, we perform a similar distribution matching on the How2Sign train and \BPCCASL\ datasets. The frame distribution before and after the frame selection is present in Fig. \ref{fig:Frame-distribution-Synthetic_WLASL-How2Sign} . Framerate matching also performed on linguistically generated dataset \BLIMPISL\ , \BLIMPASL with source dataset iSign and How2Sign. Frame Distribution in both cases is presented in Fig. \ref{fig:Frame-Distribution-Synthetic-CISLR-Blimp-iSign}, Fig. \ref{fig:Frame-Distribution-Synthetic-WLASL-Blimp-How2Sign}.  On top of that, we randomly sampled the frames with frequency ranging from [1,3] to support generalization, as the videos were from various sources. Some were from the News channel ISH, and some were educational, making the speed of the signers vary significantly. We pick the selected keypoints from the selected frames and concatenate them to form a single vector of dimension 152 fed into our Encoder model.

\subsection{Pose Processing}
\label{app:pose_stichingB2}
From the iSign and How2sign dataset videos, we extract the pose files using the media pipe library \cite{MediaPipe}. The media pipe library gives 576 key points as features. Sign language mainly involves manual(hand gestures) and non-manual (facial expression) markers. So, we focused on the upper body, especially the hands, and our facial expressions(eyebrows, lips, etc.). We handpicked some key points inspired by \cite{lin2023glossfreeendtoendsignlanguage}, which play a key role in recognizing the sign appropriately. Mediapipe extracts 576 key points, out of which 21 are for the left hand, 21 are for the right hand, 33 are for the body pose, and 468 are for the face.

We took all the key points for hands as they are one of the most important modalities for sign language. Some of the pose landmarks were not as important for sign language translation, so we ignored them.
We excluded the following pose keypoints: Rknee, Rankle, Rheel, Rfootindex, Lknee, Lankle, Lheel, Lfootindex, Leye(in), Leye(out), Reye(in), Reye(out), Mouth(2 keypoints), Lpinky, Rpinky, Lindex, Rindex, Lthumb, Rthumb, LHip, RHip. The respective keypoints are as follows: (26, 28, 30, 32, 25, 27, 29, 31, 1, 3, 4, 6, 9, 10, 17, 18, 19, 20,  21, 22, 23, 24). This leaves us with 11 key points out of the 33.
For the face, we only took the following key points. Mouthright (61), Mouthleft (291), LipsLowerOuter (17), LipsUpperOuter(0), RightEyebrowUpper(70, 105, 107), LeftEyebrowUpper (300, 334, 336), RightEyeUpper (161,158), RightEyeLower (33, 163, 153, 133), LeftEyeUpper (388, 385), LeftEyeLower (263,390,380,362), Nosetop(9). A total of 23 key points for the face. We use a total of 76 key points, and each key point has a corresponding (x,y) coordinate, which makes it 152 key points per frame.
Mediapipe gives the confidence of each key point that it predicts some of the key points were classified with very low confidence, we chose a threshold of 0.8, and if any key point was less than this threshold, then we filled the key points with the nearest (left/right) frame of the video with the confidence of the keypoints surpassing the threshold.




\section{Hyperparameters/Training Details} \label{app:section-2}


Table \ref{tab:hyperparams} shows the hyperparameters used to train the architectures on different datasets. Some of the hyperparameters differ across the datasets, such as learning rate, number of encoder and decoder layers, attention heads, and dropout.

The sampling steps indicate the step at which we sample from the generated dataset. For example, a sampling rate of 3 indicates that we sample every 3rd frame in the generated dataset. The generated dataset has a larger number of frames as the frames are stitched together and are fetched from a vocabulary dataset, so the speed at which they are signed is also slow, effectively making the number of frames large in the generated dataset. This sampling rate is derived by making the mean of the frames of the dataset (iSign / How2Sign) similar to the mean of the frames of the generated dataset. So, if the mean of the number of frames of the generated dataset is $x$ and the mean of the frames of the available dataset is $y$, then the sampling rate is equal to $x/y$.

For training, we follow two strategies: one where, while stitching the poses, the order of the words in the generated sentences is kept the same (same word order stitching (SWO)), and the second in which the order of the words in the generated sentences is shuffled randomly (random word order stitching (RWO)).
We found this simple strategy of linearly moving towards the target distribution to work well across datasets.


\subsection{Architecture Details}
\label{app:architecturalC1}
We follow an encoder-decoder model architecture for training an end-to-end SLT system. We took the BERT model \cite{devlin-etal-2019-bert} as our encoder and the GPT2 model \cite{radford2019language} as our decoder. We use the Huggingface transformers library \cite{wolf-etal-2020-transformers} for the implementation of the models. For the iSign dataset, both the encoder and the decoder had 4 layers with a hidden size of 512 with 8 attention heads each. We take the processed features/poses and feed them directly to our BERT model. For the decoder, we trained a BPE tokenizer \cite{sennrich-etal-2016-neural} with our train data from both the generated data ( \BPCCISL, \BLIMPISL ) and the iSign dataset with a vocab of 15000. We took the ADAMW \cite{loshchilov2019decoupledweightdecayregularization} as our optimizer with a learning rate of 3e-4 and a batch size of 16. We used 0.1 as our dropout. For a full set of hyperparameters, kindly refer to Table \ref{tab:hyperparams}.
For the pretraining strategy, we linearly increased the iSign data for training the model with a max threshold of 85 percent at 60000 steps. That is, we sample from a uniform random variable. If the sample is less than the threshold, we sample from iSign, or else we sample from the generated dataset. So, at the 0th step, we always sample from the generated dataset while linearly increasing the threshold to a maximum of 0.85. That is, after 60k steps, we sample from iSign 85 percent of the time. We follow a similar strategy for the How2Sign dataset with a few changes in hyperparameters.
We followed two pretraining strategies. In the first strategy, we stitched the words/poses of the sentence in the same order, while in the second, the poses were stitched in random order. Table \ref{tab:merged_results} shows the results for different pretraining.

\subsection{Linear Annealing} 
\label{app:architecturalC2}
\begin{figure}[h]
  \centering
  {
  \includegraphics[width=\linewidth]{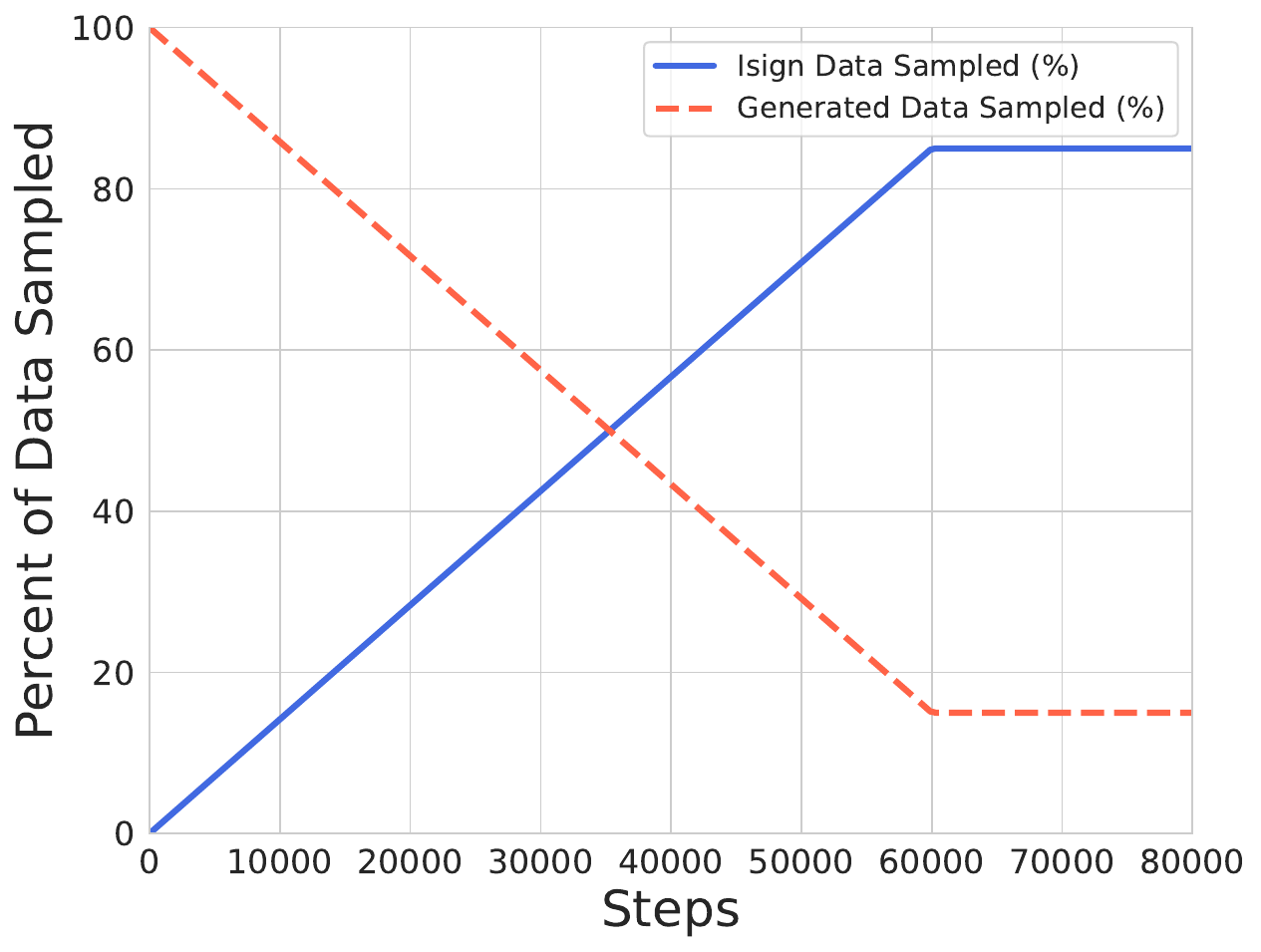}
  }
  \caption{Data sampling over the training steps}
  \label{fig:isign-complementary-sampling}
\end{figure}

During training, we follow a linear annealing strategy to first start from the constructed pose-stitched datasets and finally move towards the target distribution sentences from the respective sign language. More specifically,
we linearly increase the number of samples that we take from the original dataset and reduce the samples that we take from the generated dataset. Fig. \ref{fig:isign-complementary-sampling} shows the percent of data that we sample from the original dataset and the generated dataset as the number of training steps increases. Basically, we linearly increase the data from 0\% (iSign/How2Sign) data at the 0th step to 85\% of the (iSign/How2Sign) data at the 60000th step.
\begin{figure}[t]
  \centering
  {
  \includegraphics[width=0.75\linewidth]{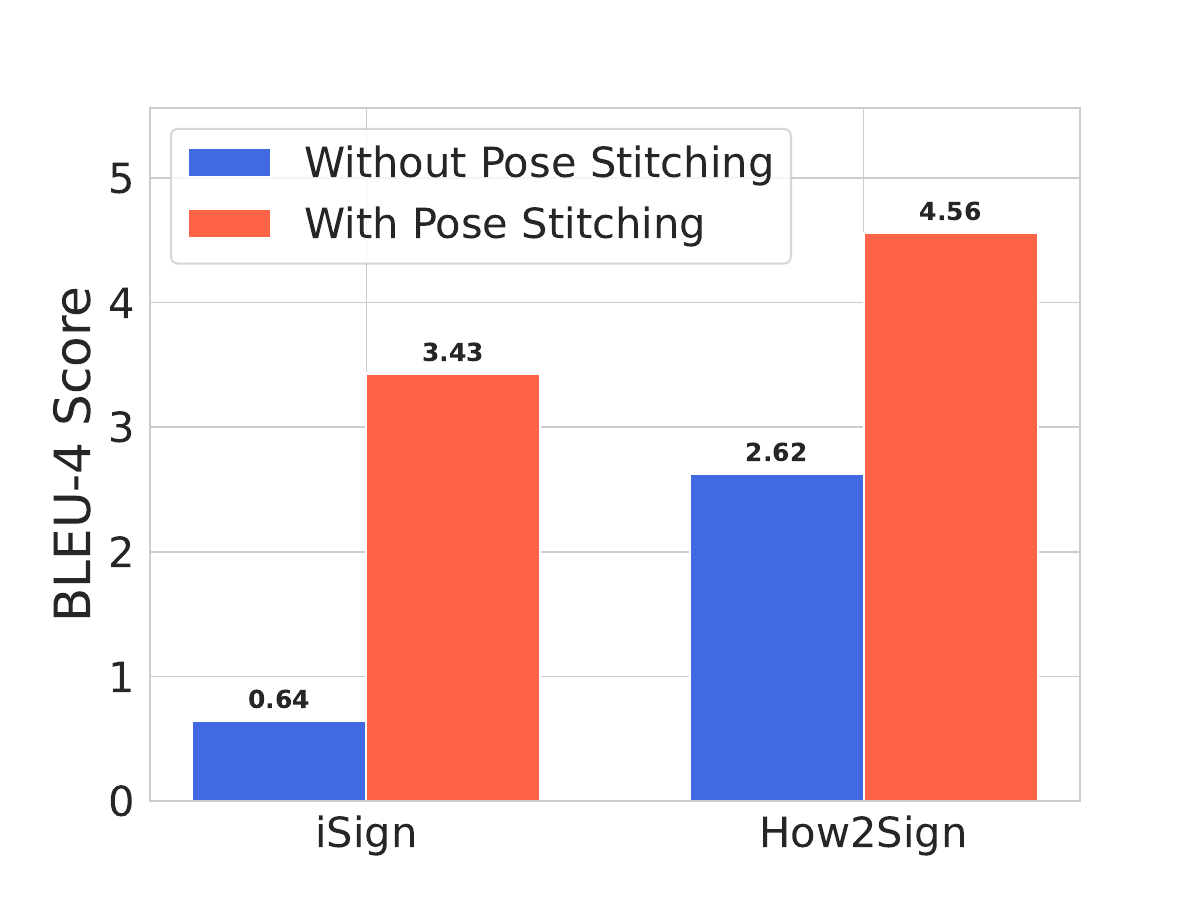}
  }
  \caption[BLEU-4 with and without pose stitching]{Performance gain from the proposed training strategy with added template-based generated sentences.}
  \label{fig:Before-after-PT}
  \vspace{-4mm}
\end{figure}
\section{Additional Results}
\label{app: Additional Results}

\subsection{Effect of Adding Pose-stitched Dataset}
\label{app: Additional ResultsD1}
\begin{figure}[h]
  \centering
  \includegraphics[width=\linewidth]{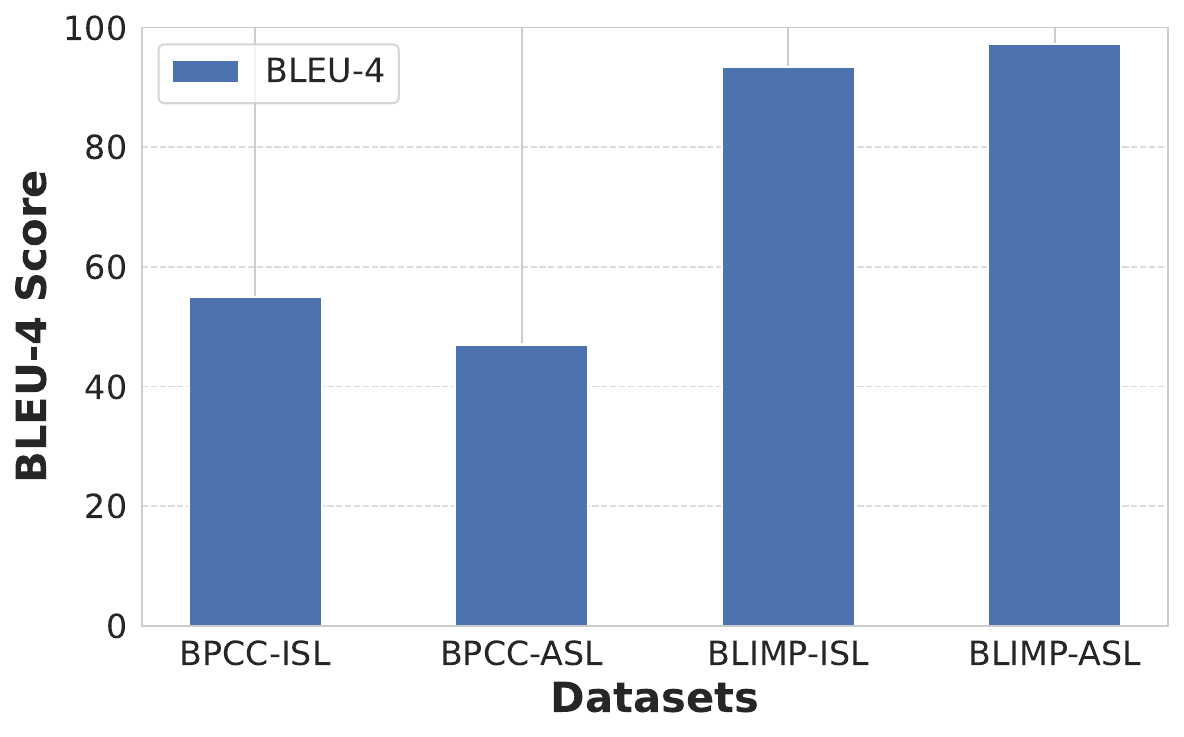}
  \caption{Validation BLEU scores on generated dataset}
  \label{fig:bleu_scores_pretraining}
\end{figure}

To further analyze whether training on the generated dataset provides any performance gain, we compared the performance of the system without training on the generated dataset to the performance after training on it. App. Table \ref{tab:no_pretraining_finetuning} shows a significant performance gain when the system is trained on the generated dataset, indicating that the proposed strategy is indeed helpful in improving the system's performance. We also ran an experiment with the same set of hyperparameters without adding any pose-stitched sentences to the training set. Fig. \ref{fig:Before-after-PT} highlights the performance boost obtained in both datasets. Further, we also measure the performance of these models on new pose-stitched sentences not seen during training. Fig. \ref{fig:bleu_scores_pretraining} shows the obtained translation scores. We observe that the model does generalize over the created pose-stitched sentences, with a BLEU-4 score of 97 and 47 for different sets. We speculate that the primary reason for this boost is the enormous dataset size with less vocabulary, making it easier to learn a generalized representation. Moreover, the distribution of the generated sentences coming from the same set of vocabulary helps generalize better. A few sample translated pose-stitched sentences with ground truth sentences are shown in App. Table \ref{tab:BLIMP-qualitative}. \\

\subsection{Effect of Distribution Shifts}
\label{app: Additional ResultsD2}
To see if our model is learning the concepts present in the training data, we took the most similar and the least similar sentences of the train split to the test and val splits of the iSign dataset and computed the BLEU score on these. Similarly, we took the most similar and the least similar sentences from the generated dataset ( \BPCCISL\ ) with the test and val splits of the dataset. We use SBERT embeddings to find the sentences with higher similarity and create two subsets with the highest and lowest similarity. Table \ref{tab:most-least-similar} indicates a positive signal for the performance on the most similar sentences, while it performs worse on the least similar sentences, indicating that the model is able to learn the context. Moreover, this pattern repeats in the generated dataset, which indicates that if we increase the dataset such that the overlap with the original dataset is high, the performance of the system can significantly improve.  Overall, we observe a performance improvement if the distribution of the sentences matches with the distribution of created pose-stitched sentences, showing the effectiveness of the proposed training strategy.
\\

\subsection{Qualitative Analysis}
\label{app: Additional ResultsD3}
For the qualitative analysis, we included a few translated sentences and their corresponding ground truth sentences from the How2Sign dataset in Table \ref{tab:ref-pred-table-How2sign}. The Table shows the performance of our Model compared to GloFE. Our Model often captures the main idea better — for example, it correctly says “That’s a good question”, while GloFE gives unrelated or incorrect outputs. In some cases, like “It’s going to blend it in the hair”, our model is partly correct, showing better alignment with the reference. However, there are also hallucinations, like in “That’s called sympathetic magic”, our model repeats “magic magic magic”, which is incorrect. Overall, our model produces more relevant and accurate translations than GloFE.

\subsection{Effects of random word order in the pretraining dataset.}
\label{app: Additional ResultsD4}
Most sign languages have a distinct and often non-linear grammatical structure that differs significantly from English. However, due to the lack of accessible linguistic resources, reliable syntactic parsers, or annotated corpora for ASL or ISL grammar, we adopt English word order as a proxy, which simplifies generation and leverages the available textual infrastructure.

\noindent To explore how sensitive the model is to this assumption, we construct two variants of our synthetic datasets: \textbf{1) \textit{Same Word Order (SWO):}} Poses are stitched in the same order as the English sentence, preserving syntactic structure and compositional cues. \textbf{2) \textit{Random Word Order (RWO):}} Poses are stitched after randomly permuting the word order, injecting syntactic noise, and encouraging the model to learn flexible and robust representations. These two variants allow us to investigate the trade-off between syntactic alignment and generalization, especially in low-resource or cross-lingual sign language translation settings.
For both How2Sign and iSign, we see that the models trained in the SWO fashion result in better performance. However, for How2Sign, RWO also comes close to the best model, indicating the model can robustly learn the representations. 


\begin{table*}[t]
\centering
\small
    \resizebox{\textwidth}{!}{
    \begin{tabular}{l cccc | cccc}
        \toprule
        \textbf{Method} & \multicolumn{4}{c}{\textbf{DEV}} & \multicolumn{4}{c}{\textbf{TEST}} \\
        \cmidrule(lr){2-5} \cmidrule(lr){6-9}
        & BLEU-1 & BLEU-2 & BLEU-3 & BLEU-4 & BLEU-1 & BLEU-2 & BLEU-3 & BLEU-4 \\
        \midrule
        Most Similar Sentences (\BPCCISL) &20.73  &10.08  &6.31  &4.43 &19.33  &9.20  &5.75  &4.00    \\
        Least Similar Sentences (\BPCCISL) &17.12  &7.66  &4.61  &3.18 &16.51  &7.37  &4.33  &2.88  \\
        Most Similar Sentences(iSign train) &19.84 &10.18 &6.74 &4.97 &20.48 &10.45 &6.82 &4.91 \\
        Least Similar Sentences(iSign train) &15.97 &6.66 &3.69 &2.40 &16.08 &6.65 &3.74 &2.44 \\
        
        \bottomrule
    \end{tabular}}
    \caption{Inspecting Pretraining effect on iSign dataset}
    \label{tab:most-least-similar}
\end{table*}

\begin{table*}[t]
    \centering
    \small
    \resizebox{\textwidth}{!}{
    \begin{tabular}{l cccc | cccc}
        \toprule
        \textbf{Method} & \multicolumn{4}{c}{\textbf{DEV}} & \multicolumn{4}{c}{\textbf{TEST}} \\
        \cmidrule(lr){2-5} \cmidrule(lr){6-9}
        & BLEU-1 & BLEU-2 & BLEU-3 & BLEU-4 & BLEU-1 & BLEU-2 & BLEU-3 & BLEU-4 \\
        \midrule
        No PreTraining iSign &12.85  &2.50  &1.04  &0.58  &12.81  &2.66  &1.14  &0.64  \\
        BPCC iSign (swo)  &17.31  &8.09  &5.02  &\textbf{3.54}  &17.67  &8.20  &5.00  &\textbf{3.43}  \\
        No Pretraining How2Sign &22.19  &9.71  &5.43  &3.37  &18.16  &7.96  &4.33  &2.62  \\
        BLIMP How2Sign (swo) &26.89  &13.36  &7.92  &\textbf{5.04}  &25.98  &12.55  &7.25  &\textbf{4.56}  \\
        
        \bottomrule
    \end{tabular}}
    \caption{Performance gain from Pretraining strategy}
    \label{tab:no_pretraining_finetuning}
\end{table*}


\begin{table}[t]
    \centering
    \resizebox{\columnwidth}{!}{
    \begin{tabular}{l c}
        \toprule
        \textbf{Hyperparameter} & \textbf{Value(Isign, How2Sign)} \\
        \midrule
        Learning Rate &  3e-4, 1e-4 \\  
        Number of Encoder Layers & 4, 2 \\
        Number of Decoder Layers & 4, 2 \\
        Encoder Hidden Size & 512 \\
        Decoder Hidden Size & 512 \\
        Number of Attention Heads & 8, 4 \\
        Dropout & 0.1, 0.3 \\
        Max Frames (Truncate after these) & 300 \\
        Number of Beams & 3 \\
        Warmup Steps Ratio & 0.1 \\
        Batch Size & 16 \\
        LR Scheduler Type & constant\_schedule\_with\_warmup \\
        Max Length Decoder & 128 \\
        Vocabulary Size Decoder & 15000 \\
        Number of Keypoints & 152 \\
        Weight Decay & 0.01 \\
        Sampling Steps(x) & 4, 3 \\
        \bottomrule
    \end{tabular}}
    \caption{Set of hyperparameters used in the experiment.}
    \label{tab:hyperparams}
\end{table}

\begin{table*}[ht!]
    \centering
    \small
    \begin{tabular}{@{}p{0.28\linewidth} p{0.32\linewidth} p{0.32\linewidth}@{}}
        \toprule
        \textbf{Reference} & \textbf{GloFE} & \textbf{Our Model} \\
        \midrule
        Hi ! & \textcolor{blue}{Hi!} & \textcolor{blue}{Hi} ! \\[4pt]
        
        In this clip we are going to talk about dangers for these birds in the household and otherwise. &
        \textcolor{red}{and i am going to show you what i want to do is you don't want it to do with it.} &
        \textcolor{blue}{In this clip, we're going to talk about the bird} \textcolor{red}{} \\[4pt]
        
        Well that's a good question. &
        \textcolor{red}{it is not} \textcolor{blue}{good} \textcolor{red}{for you.} &
        \textcolor{blue}{That’s a good question.} \\[4pt]
        
        Cross over your arms to keep the bar steady and to hold there. &
        \textcolor{red}{if you want to keep your feet and} \textcolor{blue}{keep} \textcolor{red}{your tape.} &
        \textcolor{red}{It’s very important to keep} \textcolor{blue}{your arms} \textcolor{red}{straight and it’s not much quicker.} \\[4pt]
        
        That’s called sympathetic magic. &
        \textcolor{red}{you don't want to go too much.} &
        \textcolor{red}{It’s called a} \textcolor{blue}{magic} \textcolor{red}{magic magic.} \\[4pt]
        
        Its really easy to use. &
        \textcolor{red}{so, i'm going to show you how to do this.} &
        \textcolor{blue}{It’s easy to} \textcolor{red}{do.} \\[4pt]
        
        Here we go. &
        \textcolor{red}{it doesn't have any way.} &
        \textcolor{red}{So} \textcolor{blue}{we go}. \\[4pt]
        
        Take a deep breath in through your nose. &
        \textcolor{red}{so, you're going to want to get a little bit of a little bit.} &
        \textcolor{red}{Inhale and exhale.} \\[4pt]
        
        And you would paint it on your hair. &
        \textcolor{red}{this is going to be the top of your head.} &
        \textcolor{red}{It’s going to blend it in the} \textcolor{blue}{hair}. \\[4pt]
        
        And I’m going to take this one and extend it a little bit more. &
        \textcolor{red}{the first thing you need to do is make sure you keep your hands.} &
        \textcolor{red}{So, you want to make} \textcolor{blue}{a little bit more}. \\[4pt]
        
        Can you swing your legs around? &
        \textcolor{red}{you want to make sure that you get your feet.} &
        \textcolor{red}{You can go} \textcolor{blue}{around} \textcolor{red}{the floor.} \\
        \bottomrule
    \end{tabular}
    \caption[Qualitative results on How2Sign] {Qualitative results on How2Sign comparing our model predictions and GloFE predictions with references. \textcolor{red}{Red} indicates hallucinated or incorrect segments, and \textcolor{blue}{blue} indicates correct matches with the reference.}
    \label{tab:ref-pred-table-How2sign}
\end{table*}

\begin{table*}[t]
\centering
\small
\resizebox{\textwidth}{!}
{
\begin{tabular}{p{0.48\textwidth} p{0.48\textwidth}}
\toprule
\textbf{Reference} & \textbf{Prediction} \\
\midrule
Some people cure this student . & Some people cure this student . \\
Those children will dislike themselves . & Those children will dislike themselves . \\
This niece can buy this blouse . & This niece can buy this blouse . \\
People know that they climb down this ladder . & People know that they climb down this ladder . \\
Some truck might turn . & Some truck might turn . \\
These people teach that man . & These people teach that man . \\
That doctor should watch this daughter . & That daughter should watch this doctor . \\
This sweater will stretch . & This sweater will stretch . \\
Can that woman ever worry ? & That woman can worry . \\
That teacher can see that cheap hill . & That teacher can see that cheap hill . \\
\bottomrule
\end{tabular}
}
\caption[Qualitative Results on \BLIMPISL\ ]{Reference vs Prediction sentence pairs generated using BLIMP template for qualitative comparison }
\label{tab:BLIMP-qualitative}
\end{table*}

\begin{table*}[t]
    \centering
    \resizebox{\linewidth}{!}{
    \small
    \setlength{\tabcolsep}{5pt}
    \begin{tabular}{l l l ccc ccc}
        \toprule
        & \textbf{Methods} & \textbf{Source} & \multicolumn{3}{c}{\textbf{DEV}} & \multicolumn{3}{c}{\textbf{TEST}} \\
        \cmidrule(lr){4-6} \cmidrule(lr){7-9}
        & & & ROUGE-1 & ROUGE-2 & ROUGE-L & ROUGE-1 & ROUGE-2 & ROUGE-L \\
        \midrule
        \multirow{7}{*}{\rotatebox{90}{How2Sign(ASL)}} 
        & GloFE-VN & -- & 15.48 & 3.40 & 13.36 & 15.39 & 3.52 & 13.27 \\
        \cmidrule(lr){2-9}
        & w/o Pose Stitched & -- & 17.76 & 4.38 & 14.85 & 16.55 & 3.67 & 13.98 \\
        & \multirow{2}{*}{Pose Stitched (rwo)}
        & BPCC & 21.71 & 6.00 & \textbf{18.28} & 20.72 & 5.61 & 17.53 \\
        & & BLIMP & 20.55 & 5.86 & 17.34 & 19.90 & 5.23 & 16.83 \\
        & \multirow{2}{*}{ Pose Stitched (swo)} 
        & BPCC & 21.27 & 5.67 & 18.04 & \textbf{21.23} & \textbf{5.79} & \textbf{17.94} \\
        & & BLIMP & \textbf{21.36} & \textbf{6.11} & 18.23 & 20.13 & 5.25 & 16.96 \\
        \cmidrule(lr){4-9}
        & Best & -- & \textbf{21.36 (+5.88)} & \textbf{6.11 (+2.71)} & \textbf{18.28 (+4.92)}  & \textbf{21.23 (+5.84)} & \textbf{5.79 (+2.27)} & \textbf{17.94 (+4.67)}  \\
        \midrule
        \multirow{7}{*}{\rotatebox{90}{iSign (ISL)}} 
        & GloFE-VN & -- & 10.26 & 1.32 & 9.26 & 9.72 & 1.32 & 8.88 \\
        \cmidrule(lr){2-9}
        & w/o Pose Stitched & -- & 9.23 & 0.53 & 8.25 & 1.49 & 0.40 & 1.45 \\
        & \multirow{2}{*}{Pose Stitched (rwo)} & BPCC & 13.31 & 3.06 & 11.81 & 13.44 & 3.15 & 11.86 \\
        & & BLIMP & 11.15 & 1.87 & 09.95 & 11.72 & 2.06 & 10.41 \\
        & \multirow{2}{*}{Pose Stitched (swo)} & BPCC & \textbf{16.00} & \textbf{4.13} & \textbf{14.14} & \textbf{16.37} & \textbf{4.28} & \textbf{14.41} \\
        & & BLIMP & 15.54 & 3.98 & 13.88 & 15.59 & 4.01 & 13.92 \\
        \cmidrule(lr){4-9}
        & Best & -- & \textbf{16.00 (+5.74)}  & \textbf{4.13 (+2.81)}  & \textbf{14.14 (+4.88)} & \textbf{16.37 (+6.65)}  & \textbf{4.28 (+2.96)}  & \textbf{14.41 (+5.53)}  \\

        \bottomrule
    \end{tabular}}
    \caption[ROUGE score result]{ROUGE score results on the How2Sign and iSign datasets comparing different pretraining strategies (random word order vs. same word order), along with baseline results (GloFe). The numbers in brackets show the absolute improvements from the baseline.}
    \vspace{-5mm}
    \label{tab:rouge_results}
\end{table*}

\begin{table*}[t]
    \centering
    \resizebox{\linewidth}{!}{
    \small
    \setlength{\tabcolsep}{5pt}
    \begin{tabular}{l l l cccc cccc}
        \toprule
        \textbf{Dataset} & \textbf{Methods} & \textbf{Source} & \multicolumn{4}{c}{\textbf{DEV}} & \multicolumn{4}{c}{\textbf{TEST}} \\
        \cmidrule(lr){4-7} \cmidrule(lr){8-11}
        & & & SacreBLEU1 & SacreBLEU2 & SacreBLEU3 & SacreBLEU4 & SacreBLEU1 & SacreBLEU2 & SacreBLEU3 & SacreBLEU4 \\
        \midrule
        How2Sign 
        & Pose Stitched (swo) 
        & BLIMP & 26.88 & 6.63 & \textbf{2.79} & \textbf{1.29} & 25.97 & 6.06 & 2.41 & \textbf{1.13} \\
        \midrule
        iSign 
        & Pose Stitched (swo) 
        & BLIMP & \textbf{21.99} & 4.59 & 2.34 & 1.48 & \textbf{22.36} & 4.52 & 2.25 & \textbf{1.37} \\
        \bottomrule
    \end{tabular}}
    \caption[Sacre score result]{Best SacreBLEU scores on the How2Sign and iSign datasets comparing different pretraining strategies (same word order).}
    \label{tab:sacre_results}
\end{table*}

\end{document}